\documentclass[runningheads]{llncs}

\usepackage{cite}
\usepackage{amsmath,amssymb,amsfonts}
\usepackage{subfiles}
\usepackage{setspace}
\usepackage{graphicx}
\usepackage{textcomp}
\usepackage{xcolor}
\usepackage{times}

\usepackage{amsmath}
\usepackage{amssymb}
\usepackage{wrapfig, booktabs}
\newcommand{\R}{\mathbb{R}}
\usepackage{soul}
\usepackage{url}
\usepackage[hidelinks]{hyperref}
\usepackage[utf8]{inputenc}
\usepackage[small]{caption}
\usepackage{graphicx}
\usepackage{float}
\restylefloat{table}
\usepackage[tableposition=top]{caption}
\usepackage{float}
\floatstyle{plaintop}
\restylefloat{table}
\usepackage{subcaption}
\usepackage{mwe}
\usepackage{amsmath}
\usepackage{booktabs}
\usepackage{algorithm}
\usepackage{multirow}

\usepackage{algpseudocode}
\usepackage{adjustbox}

\usepackage{amsthm}
\usepackage{wrapfig}

\usepackage{resizegather}
\usepackage{stackengine}
\usepackage[misc]{ifsym}
\def\delequal{\mathrel{\ensurestackMath{\stackon[1pt]{=}{\scriptstyle\Delta}}}}

\theoremstyle{definition}

\begin{document}
\author{Sumaiya Tabassum Nimi(\Letter) \and
Md Adnan Arefeen \and
Md Yusuf Sarwar Uddin \and Yugyung Lee}
\authorrunning{Sumaiya et al.}
\tocauthor{Sumaiya Tabassum Nimi, Md Adnan Arefeen, Md Yusuf Sarwar Uddin, and Yugyung Lee}
\institute{University of Missouri-Kansas City, MO, USA \\
\email{\{snvb8,aa4cy\}@mail.umkc.edu,\{muddin,LeeYu\}@umkc.edu}}
\title{EARLIN: Early Out-of-Distribution Detection for Resource-efficient Collaborative Inference}
\toctitle{EARLIN: Early Out-of-Distribution Detection for Resource-efficient Collaborative Inference}
%~\thanks{To appear in ECML-PKDD'2021.}}
\titlerunning{EARLIN}

\maketitle

\begin{abstract}
%{\bf this abstract seems too long. The length of about 300 words is good.}
Collaborative inference enables resource-constrained edge devices to make inferences by uploading inputs (e.g., images) to a server (i.e., cloud) where the heavy deep learning models run. While this setup works cost-effectively for successful inferences, it severely underperforms when the model faces input samples on which the model was not trained (known as Out-of-Distribution (OOD) samples). If the edge devices could, at least, detect that an input sample is an OOD, that could potentially save communication and computation resources by not uploading those inputs to the server for inference workload. In this paper, we propose a novel lightweight OOD detection approach that mines important features from the \emph{shallow} layers of a pretrained CNN model and detects an input sample as ID (In-Distribution) or OOD based on a distance function defined on the reduced feature space. Our technique (a) works on pretrained models without any retraining of those models, and (b) does not expose itself to any OOD dataset (all detection parameters are obtained from the ID training dataset). To this end, we develop EARLIN (\textbf{EARL}y OOD detection for Collaborative \textbf{IN}ference) that takes a pretrained model and partitions the model at the OOD detection layer and deploys the considerably small OOD part on an edge device and the rest on the cloud. By experimenting using real datasets and a prototype implementation, we show that our technique achieves better results than other approaches in terms of overall accuracy and cost when tested against popular OOD datasets on top of popular deep learning models pretrained on benchmark datasets.

\end{abstract}


%section
\section{Introduction}
With the emergence of Artificial Intelligence (AI), applications and services using deep learning models, especially Convolutional Neural Networks (CNN), for performing intelligent tasks, such as image classification, have become prevalent. However, several issues have been observed in deployment of the deep learning models for real-life application. First, since the models tend to be very large in size (100's of MB in many cases), they require higher computation, memory, and storage to run, which makes it difficult to deploy them on end-user/edge devices. Second, these models usually predict with high confidence, even for those input samples that are supposed to be unknown to the models (called out-of-distribution (OOD) samples)~\cite{nguyen2015deep,szegedy2013intriguing}. Since both in-distribution (ID) and OOD input samples are likely to appear in real-life settings, OOD detection has emerged as a challenging research problem.
%Despite reporting spectacular results, several issues have been observed when these models are to be deployed on end-user devices or edge-devices in real-life applications. 
%First, these models are very large in sizes (100's of MB in many cases), which leads to higher computation, memory, and storage requirements to host and run them. As a result, this is challenging to deploy these models in their full sizes on edge devices and do inferences using them. 
%Second, these models usually tend to give a high confidence prediction, even when the input image does not belong to its domain of training data~\cite{nguyen2015deep,szegedy2013intriguing}. These input samples are usually referred to as out-of-distribution (OOD) samples, whereas the similar looking images to the training dataset are called in-distribution (ID) samples. Since real-life inputs probably could be a mix of both ID and OOD data, the problem of detecting these OOD samples before classification has emerged as a challenging field of research in recent years. Popular deep learning models trained on a certain dataset do not usually have the capability of detecting OOD samples in-built. 

The first issue, the deployment of deep learning models in end/edge devices, has been studied well in the literature~\cite{ laskaridis2020spinn}. One solution is to run collaborative inference, in which the end devices do not run the heavy model on-board, instead offload the inference task by uploading the input to a nearby server (or to the cloud in appropriate cases) and obtain the inference/prediction results from there. Other recent works propose doing edge-cloud collaboration~\cite{gazzaz2019collaborative}, model compression~\cite{schindler2018towards} or model splitting~\cite{kang2017neurosurgeon} for faster inference.
The second issue, the OOD detection, has received much attention in the deep learning research community~\cite{cardoso2017weightless,xie2019slsgd,deecke2018image,hendrycks2016baseline,liang2018enhancing}. We note several gaps in these research works, particularly their suitability of deployment in collaborative inference setup. Firstly, in most of these works, the input data were detected as an OOD sample using the outputs from the \emph{last}~\cite{liang2018enhancing,lee2020multi,mohseni2020self,hsu2020generalized} or \emph{penultimate}~\cite{lee2018simple} layer of the deep learning classifiers. We argue that detecting an input sample as OOD after these many computations are done by the model is inefficient. Secondly, most of the OOD detection approaches rely on full retraining the original classifier model to enable the OOD detection~\cite{mohseni2020self,lee2017training, lee2020multi}, which is computationally very expensive. Thirdly, in most of these works~\cite{liang2018enhancing,lee2018simple,mohseni2020self}, several model hyperparameters for the detection task need to be tuned based on a validation dataset of OOD samples. The fitted model is then tested, thereby inducing bias towards those datasets. Finally, some OOD detector requires computationally expensive pre-processing of the input samples~\cite{liang2018enhancing, lee2018simple}.
%If a sample could be detected as OOD earlier, it would not be a candidate for classification and thus unnecessary computation and communication can be avoided.
%The crux of the OOD detection problem lies in the fact that the set of OOD data is vast and potentially unknown. Hence such bias is unwanted and does not guarantee similar performance on unseen data. 
%Finally, computationally expensive (requires two forward and one backward passes over the classifier model) pre-processing of the input samples is needed for most of the proposed state-of-the-art OOD detectors to work~\cite{liang2017enhancing, lee2018simple}.

%\textcolor{red}{Describe Collaborative inference and distributed inference issues used as edge-cloud collaboration}

In this paper, we tackle the above two discussed issues jointly. We propose a novel OOD detection approach, particularly for Convolutional Neural Networks (CNN) models, that detects an input sample as OOD early into the network pipeline, using the portion of the feature space obtained from the shallow layers of the pretrained model. 
%deep learning image classifier for subset of training ID samples
It is documented that early layers in CNN models usually pick up some salient features representing the overall input space whereas the deeper layers progressively capture more discriminant features toward classifying the input samples to the respective classes~\cite{matthew2014visualizing}. This, therefore, suggests that these salient feature maps extracted from a designated \emph{early} layer will be different for ID and OOD samples. This is the principle observation based on which we attempt to build our OOD detector model. However, the space spanned by the obtained feature maps is in most of the cases too big to make any significant partitioning between ID and OOD samples. Hence we compress the high dimensional feature space by mining the \emph{most significant} information out of the space. We apply a series of ``feature selection'' operations on the extracted feature maps, namely \textit{indexed-pooling} and max-pooling, to reduce the large feature space to a manageable size. After the reduction, we construct a distance function defined on the reduced feature space so that the distance measure can differentiate ID and OOD samples. For deployment in edge-cloud collaboration setup, we partition the model around the selected layer to obtain a super-small OOD detection model and readily deploy the lightweight model on an edge device. With that, the edge device can detect an incoming input sample as OOD and if detected, does not upload the sample to the server/cloud (thus saves communication and computation resources).
%and thus can save valuable computation and communication resources by not offloading that the sample to the server/cloud for further inference at the expense of small processing for OOD detection. 

To this end, we develop EARLIN (EARLy OOD Detection for Collaborative INference) based on the our proposed OOD detection technique. We evaluate EARLIN on a set of popular CNN models for image classification, namely Densenet, ResNet34, ResNet44, and VGG16 models pretrained on benchmark image datasets CIFAR-10 and CIFAR-100~\cite{cifar}. We also compare our OOD detection algorithm with state-of-the-art other OOD detection techniques discussed in the literature. Furthermore, we design and develop an OOD-aware collaborative inference system and show that this setup results in faster and more precise inference in edge devices. To the best of our knowledge, ours is the first work to propose such OOD-aware collaborative inference framework. Furthermore, we define a novel performance metric, the \emph{joint accuracy} of a model combined with its detector, to quantify the performance of the model and detector combination, and formally characterize EARLIN's performance and cost using that metric.
We summarize our contributions as follows:
\begin{itemize}
    \item We propose a novel OOD detection approach called EARLIN that enables detection of OOD samples early in the computation pipeline, with minimal computation. 
    %We thereby avoid unnecessary processing in the deeper layers, leading to increased throughput during batch inference.
    \item Our technique does not require retraining the neural network classifier and thus can be implemented as an external module on top of available pretrained classifiers.
    \item We do not exploit samples from unknown set of OOD data for tuning hyperparamters, thereby reducing bias towards any subset of the unknown set of OOD samples.
   % \item No computationally expensive preprocessing of the input samples is required for our approach. Hence our approach is efficient.
    \item We propose a novel OOD-aware edge-cloud collaborative setup based on our proposed detector for precise and  resource-efficient inference at edge devices, along with characterizations of its performance and cost.
    
\end{itemize}

%section
\section{Related Work}
Deep Learning based methods have been designed to achieved huge success in recent years in recognition tasks but they have their limitations. The problem of reporting high confidence for all input samples, even those outside the domain of training data is inherent in the general construct of the popular deep learning models. In order to deploy the deep learning models in real-life applications, this issue should be mitigated. Hence in the recent years, a large number of research works have been conducted towards this direction. In~\cite{hendrycks2016baseline}, confidence of the deep learning classifiers in the form of output softmax probability for the predicted class was used to differentiate between ID and OOD samples.% Also, evaluation metrics were defined to determine the efficacy of the OOD detectors. 
 %It became evident from this work that confidence of classification was not enough to separate the OOD samples from the ID samples. 
Later, in~\cite{liang2018enhancing, lee2018simple}, OOD detection approaches were proposed that worked without making any change to the original trained deep learning models. We note several limitations in the works. Firstly, samples are detected as OOD at the very last layer of the classifier, thereby wasting computational resources on unnecessary computations done on input samples, that is eventually going to be identified unsuitable for classification. Secondly, the hyperparameters were tuned while being exposed to subset of OOD samples that the approach was tested on, inducing bias towards those samples. Also, due to this exposure, it can not be guaranteed that the approach will be as successful on any completely different set of OOD data. Thirdly, this approach required computationally heavy preprocessing of the input samples for the approach to work. The preprocessing involved two forward and one backward
passes over the classifier model, rendering the approach completely unsuitable for real-time deployment. %\newline
%In~\cite{lee2018simple}, another module for OOD detection was proposed that worked using features generated by both shallower and the deeper layers of the classifier models. Although this approach did not require retraining the classifier, it came with the limitation that it directly used subset OOD samples on which the approach was tested, to train a logistic regression model for finding the confidence score. %A measure to find confidence score was defined using Mahalanobis distance obtained through Gaussian Discriminant Analysis of the feature maps, thereby proposing an alternative to softmax function for finding class-confidence score.  Hence the previously discussed issue of bias existed in this work also. Moreover, the approach exploited features from penultimate layer of classifier also and hence the issue of unnecessary calculations also prevailed.\par
In the recent literature~\cite{lee2020multi, mohseni2020self, hsu2020generalized} also, OODs were not detected on the top of readily available deep learning models that were pretrained with traditional cross entropy loss, instead retraining was required. %Instead, the deep classifiers needed to be trained from scratch with custom defined objective function.
%Another recent work for OOD detection~\cite{mohseni2020self} also required additional training of the classifier for enabling OOD detection, other than minimizing the cross-entropy loss for classification. 
A better approach was proposed in~\cite{yu2020convolutional}, that did not require retraining the classifier.
%Instead, an OOD detection method was proposed that exploited informative patterns obtained from feature maps through pooling based on Normalized Compression Distance~\cite{li2004similarity}. 
Although this approach achieved good performance on OOD detection, we note that best reported results were obtained when feature maps from the deeper layers were used. %In the current work, we instead use feature maps from the shallow layers as we argue that detecting the OOD samples as early as possible allows us to discard a potential OOD sample and move onto the next input, other than wasting resources by doing further computations on the sample. 
\begin{table}[!tbp]
\centering
\caption{Comparison of approaches.}
\label{tab:comparison}
%\scriptsize
%\scalebox{0.7}{
\adjustbox{width=\linewidth}{%
\begin{tabular}{l|c|c|c|c|c|c}\toprule[2pt]
&Baseline~\cite{hendrycks2016baseline} &ODIN~\cite{liang2018enhancing} &Mahalanobis~\cite{lee2018simple} &MALCOM~\cite{yu2020convolutional} &DeConf~\cite{hsu2020generalized} &EARLIN \\\midrule[2pt]
Without Retraining? &$\checkmark$ &$\checkmark$ &$\checkmark$ &$\checkmark$ &$\times$ &$\checkmark$ \\ \midrule[1pt]
Before Last layer? &$\times$ &$\times$ &$\times$ &$\checkmark$ &$\times$ &$\checkmark$ \\ \midrule[1pt]
Use One Layer Output? &$\checkmark$ &$\checkmark$ &$\times$ &$\times$ &$\checkmark$ &$\checkmark$ \\ \midrule[1pt]
Without OOD Exposed? &$\checkmark$ &$\times$ &$\times$ &$\checkmark$ &$\checkmark$ &$\checkmark$ \\ \midrule[1pt]
Without Input Preprocessing? &$\checkmark$ &$\times$ &$\times$ &$\checkmark$ &$\times$ &$\checkmark$\\
\bottomrule[2pt]
\end{tabular}}
%}
\end{table}

%\section
\section{Proposed OOD Detection Approach: EARLIN}
\begin{figure}[]
\centering
    \fbox{\includegraphics[width=0.9\linewidth]{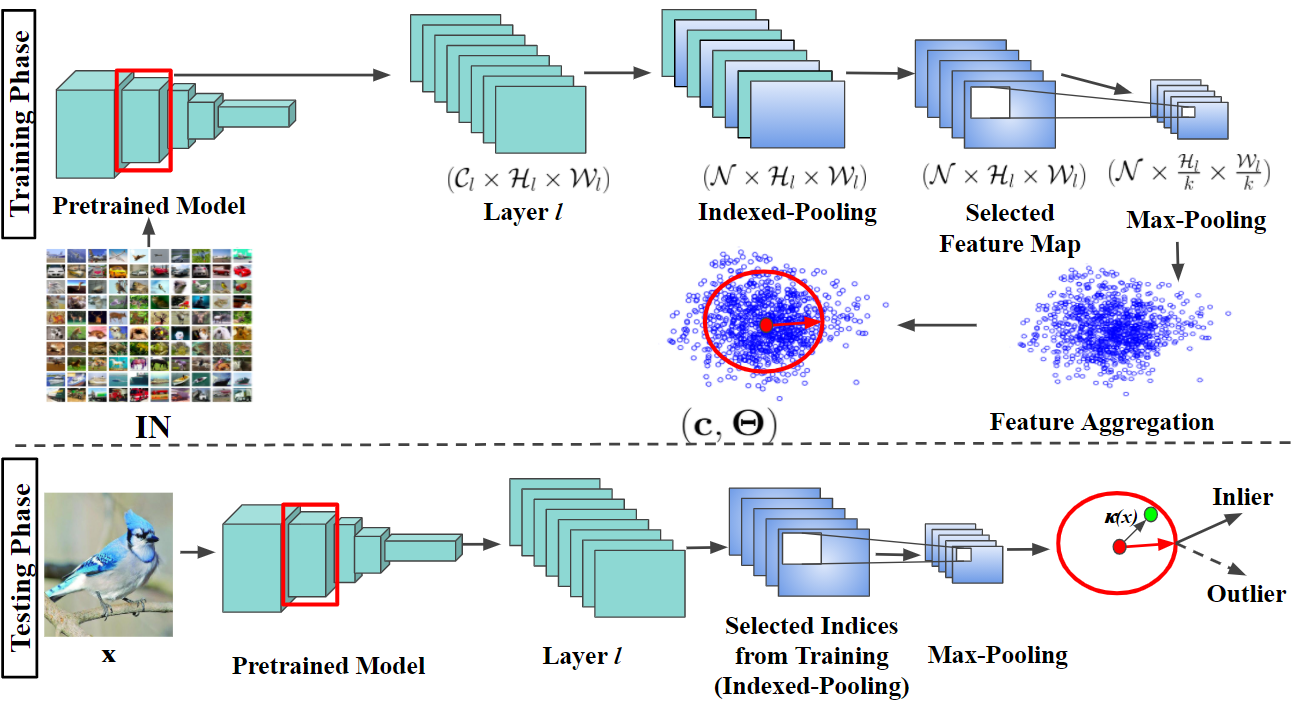}}
    \caption{Framework of the training and inference using EARLIN.}
    \label{fig:archi}

\end{figure}

We propose an OOD detection approach, called EARLIN (\textbf{EARL}y OOD detection for Collaborative \textbf{IN}ference), that enables OOD detection from the shallow layers of a neural network classifier, without requiring to retrain the classifier and without exposure to any OOD sample during training. EARLIN infers a test input sample to be ID or OOD as follows. It first feeds the input to the classification CNN model and computes up to a designated shallow layer of the model and extracts feature maps from that layer. The output of the intermediate layer is a stack of 2D feature maps, out of which a small subset of them are selected. The selected maps are ones that supposedly contain the most information entailed by all of those maps. This process is called \emph{indexed-pooling} the parameters of which (i.e., the positions of 2D maps to be selected) are determined from the training ID dataset during the training phase of the process. We then do \emph{max-pooling} for downsampling the feature space even further. With that, we obtain a vector representation of the original input in some high dimensional feature space. During training, we do this for a large number of samples drawn from the training ID dataset, aggregate them in a single cluster, and find the centroid of the ID space. Consequently, we define a distance function from the ID samples to the centroid such that at a certain level of confidence, it can be asserted that the sample is ID if the distance is less than a threshold. Since the threshold is a measure of distance, its value is expected to be low for ID samples. During inference, we use this obtained value of threshold to differentiate between ID and 
OOD samples. The framework of our proposed ID detection approach is shown in Figure~\ref{fig:archi}. 
\textbf{Feature Selection and Downsampling: Indexed-Pooling and Max-Pooling: } We at first select a subset of 2D feature maps from a designated shallow intermediate layer of pretrained neural network classifier based on a quantification of the amount of information each 2D feature map contains. We denote the chosen layer by $\ell$. From this layer, we choose the most informative $\mathcal N$ feature maps. We know from studies done previously that feature maps at shallow layers of the classifier capture useful properties out of input images~\cite{matthew2014visualizing}, but the space spanned by the feature maps is too big to capture properties inherent to the ID images out of this space. Hence we reduce the feature space by selecting a subset of the features. Visual observation reveals that some of the maps, the ones for which we obtain almost monochromatic plots, do not carry significant observation about the input image. Whereas, there are some maps that capture useful salient features from the image. We consider \emph{variance} of a 2D feature map as the quantification of the amount of information contained in the map.

\iffalse
\begin{figure}[h!t]
\centering

    \fbox{\includegraphics[width=.86\linewidth]{Images/framework_inference_v2.png}}
    \caption{ Inference using proposed OOD Detection approach.}
    \label{fig:archi}

\end{figure}
\fi

Suppose at layer $\ell$, feature map, $f \in \R^{{\cal C}_\ell \times {\cal H}_\ell \times {\cal W}_\ell}$ is obtained. So we have a total of ${\cal C}_\ell$ 2D maps (also known as channels), each of which with a dimension of ${\cal H}_\ell \times {\cal W}_\ell$. Our goal is to select $\mathcal N$ most informative 2D feature maps out of these ${\cal C}_\ell$ maps. For finding the most informative feature maps using this assumption, we choose a subset of ID training data, $\textbf D_{in}$. Using each data sample from $\textbf D_{in}$, we calculate feature map $f$, with shape  ${\cal C}_\ell \times {\cal H}_\ell \times {\cal W}_\ell$ from layer $\ell$ and finally obtain collection of feature maps, $\mathcal F$, with shape 
 $|\textbf D_{in}| \times {\cal C}_\ell \times {\cal H}_\ell \times {\cal W}_\ell$ for $\textbf D_{in}$. We define information contained in each feature map $j$, denoted as $\psi(j)$, as the summation of variance (aggregate variance) of 2D maps obtained from all input sample in the training ID dataset ($\textbf D_{in}$). This collectively measures how important map $j$ in layer $\ell$ is with respect to the entire ID population. Formally, we compute:
\begin{equation}
%\label{eq1}
    \psi_j =  \sum_{\textbf{x} \in \textbf D_{in}}{Var}({m_\ell^{\cal M}}(\textbf{x})[j]) \label{eq:psi}
\end{equation}
 where $m_\ell^{\cal M}$ denotes layer $\ell$ of model ${\cal M}$ with a tensor of size ${\cal C}_\ell \times {\cal H}_\ell \times {\cal W}_\ell$ and $m_\ell^{\cal M}[j]$ denote $j$-th 2D map in that layer having the size of size ${\cal H}_\ell \times {\cal W}_\ell$.
 Once the $\psi$ values are obtained from all ${\cal C}_\ell$ maps, we find the order statistic of $\psi$ values (sort the values in the descending order) as such:
$$\psi_{(1)} \geq \psi_{(2)} \geq \cdots \geq \psi_{(\cal N)} \geq \cdots \geq \psi_{(\cal C_\ell)}$$

We then find the \emph{indices} of top $\cal N$ channels that have the largest aggregate variance across the ID training dataset and populate a binary index vector $\mathbf{\gamma}$ to denote whether a certain map from that layer $\ell$ is selected or not. More precisely,

\begin{equation*}
%\label{eq1}
    \gamma_{j} = \begin{cases}
  1, & \text{if } \psi_j \leq \psi_{(\cal N)} \\
  0, & \text{otherwise}.
\end{cases}
\end{equation*}

Obviously, $\sum_{j=1}^{{\cal C}_\ell}\gamma_j = \cal N$. Given this binary index-vector, $\mathbf{\gamma}$, and the layer output of $m_\ell^{\cal M}(\textbf{x})$ for an input sample $\textbf{x}$, the \emph{indexed-pooling} operation takes out only those feature maps (channels) as specified by the index-vector thus effectively reduces the feature space dimension from ${\cal C}_\ell \times {\cal H}_\ell \times {\cal W}_\ell$ to ${\cal N} \times {\cal H}_\ell \times {\cal W}_\ell$. Consequently, we define the indexed-pooling operator as $\Gamma_1:\mathbb{R}^{{\cal C}_\ell \times {\cal H}_\ell \times {\cal W}_\ell} \rightarrow \mathbb{R}^{{\cal N} \times {\cal H}_\ell \times {\cal W}_\ell}$ as follows:
\begin{equation}
 \Gamma_1(\textbf{x}) = \mathbin\Vert_{j=1}^{{\cal C}_\ell} m_\ell^{\cal M}(\textbf{x})[j] \text{ if } \gamma_j = 1
\end{equation}
\noindent
where $\mathbin\Vert$ indicates concatenation.

%\subsection{Downsampling Feature Space by Pooling}
We note that the feature space spanned by the chosen $\mathcal N$ feature maps from layer $\ell$ is still too large to capture useful information. Hence, we downsample the space by $(k,k)$ max-pooling. Max-pooling is an operation that is traditionally done within the deep leaning model architectures for downsampling the feature space, so that only the most relevant information out of a bunch of neighboring values is retained. We follow the same practice for downsampling our feature space. The max-pooling operator is denoted as $\Gamma_2:\mathbb{R}^{{\cal N} \times {\cal H}_\ell \times {\cal W}_\ell}~\rightarrow~\mathbb{R}^{{\cal N} \times \frac{{\cal H}_\ell}{k} \times \frac{{\cal W}_\ell}{k}}$. We  usually use $k = 4$. 
%For the candidate CNN models we consider, we use $k = 4$. 

\begin{figure*}[!tpb]
    \centering
\scalebox{1}{
    \begin{subfigure}{0.24\textwidth}
        \centering
        \includegraphics[width=\textwidth]{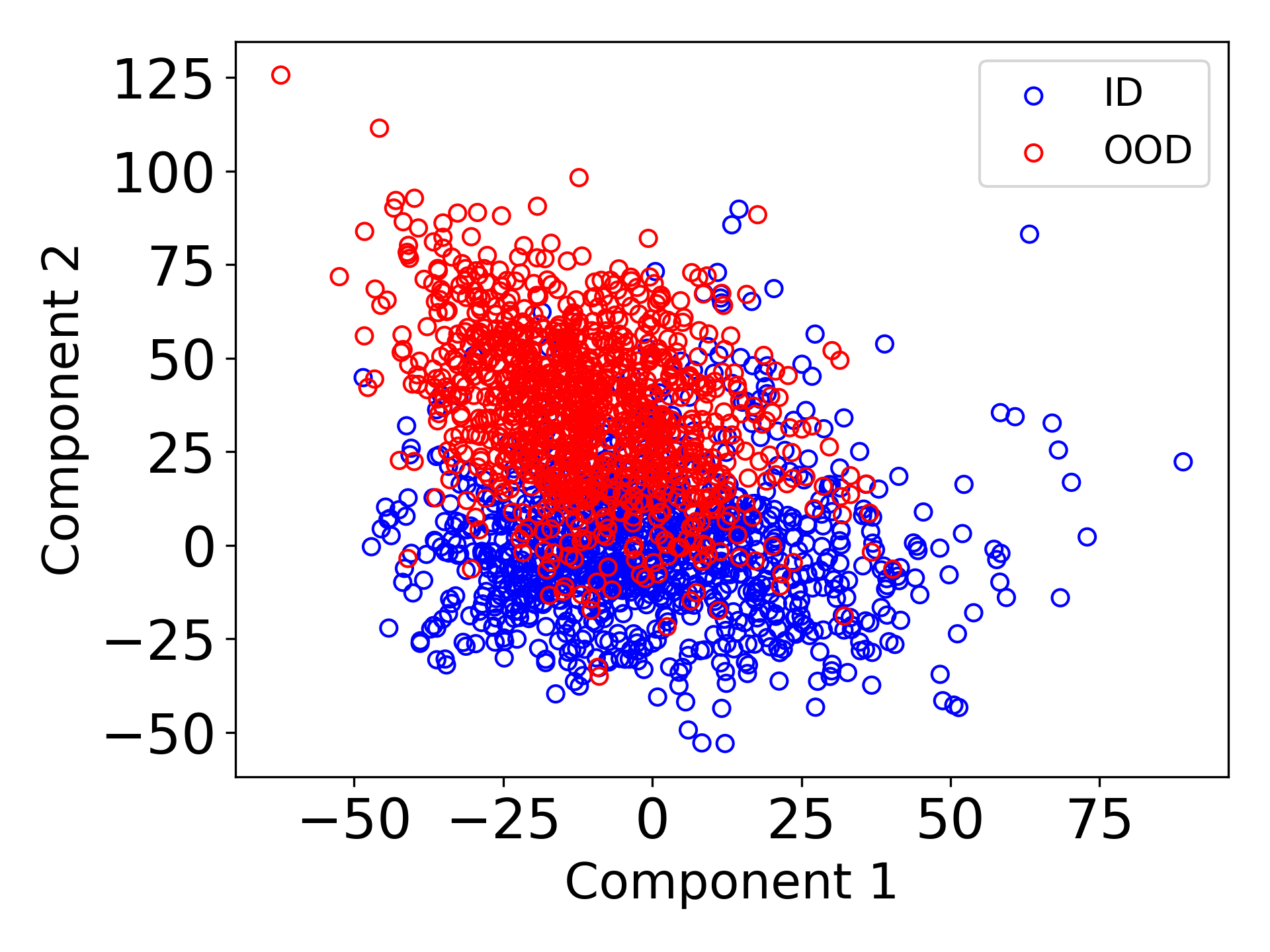}
        \caption{}
    \end{subfigure}%
    ~ 
    \begin{subfigure}{0.24\textwidth}
        \centering
        \includegraphics[width=\textwidth]{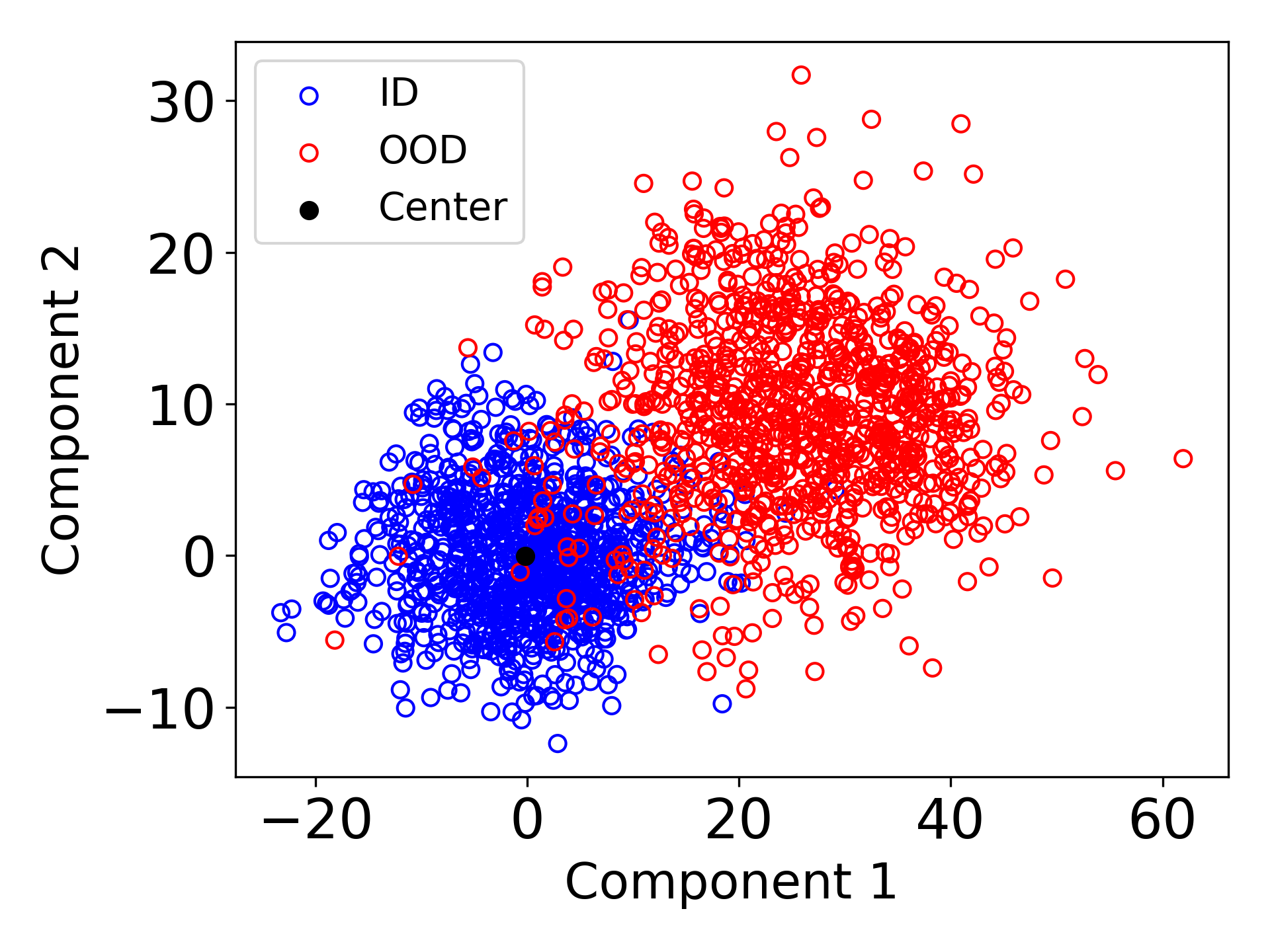}
        \caption{}
    \end{subfigure}
     \begin{subfigure}{0.24\textwidth}
        \centering
        \includegraphics[width=\textwidth]{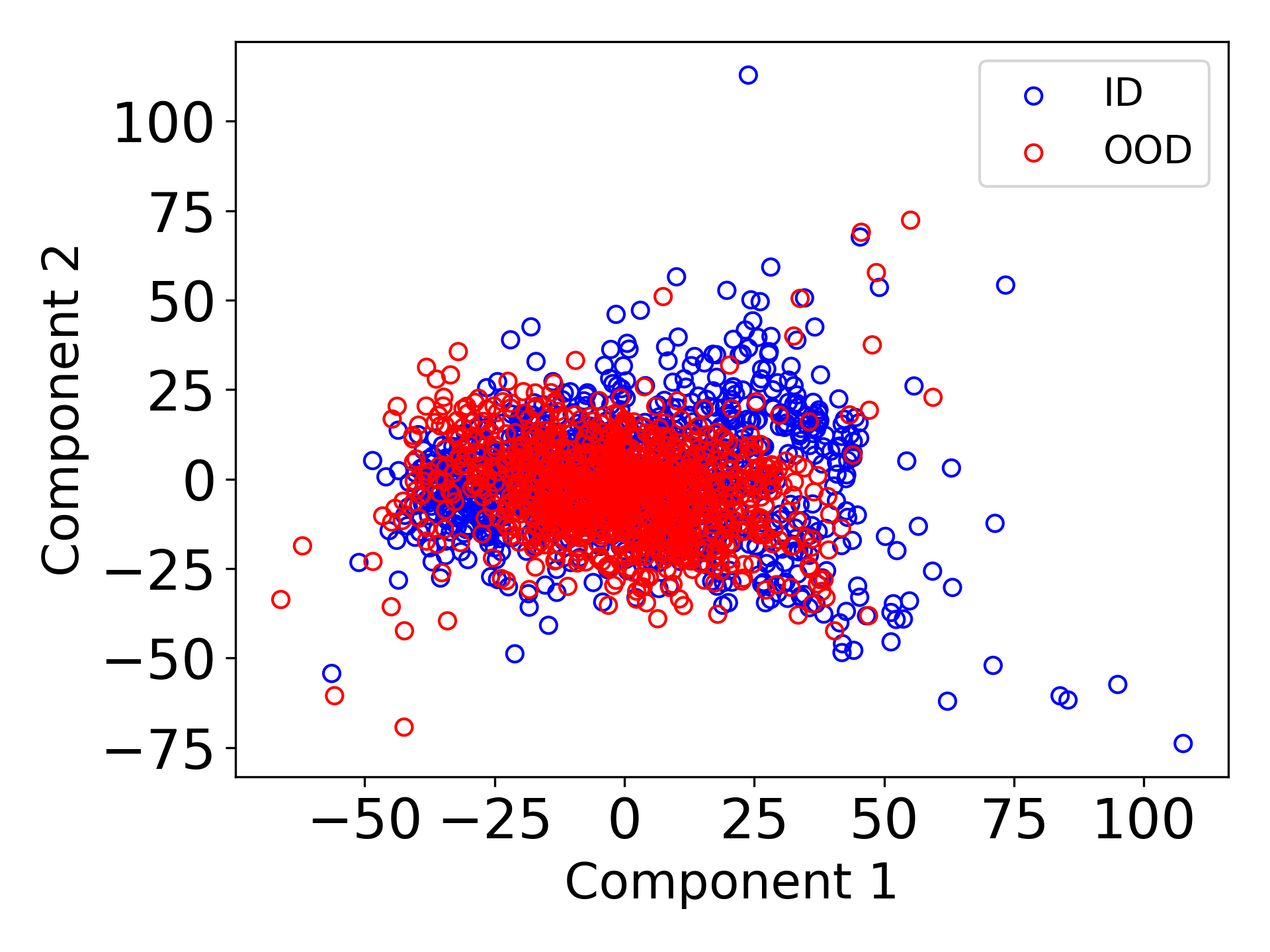}
        \caption{}
    \end{subfigure}
     \begin{subfigure}{0.24\textwidth}
        \centering
        \includegraphics[width=\textwidth]{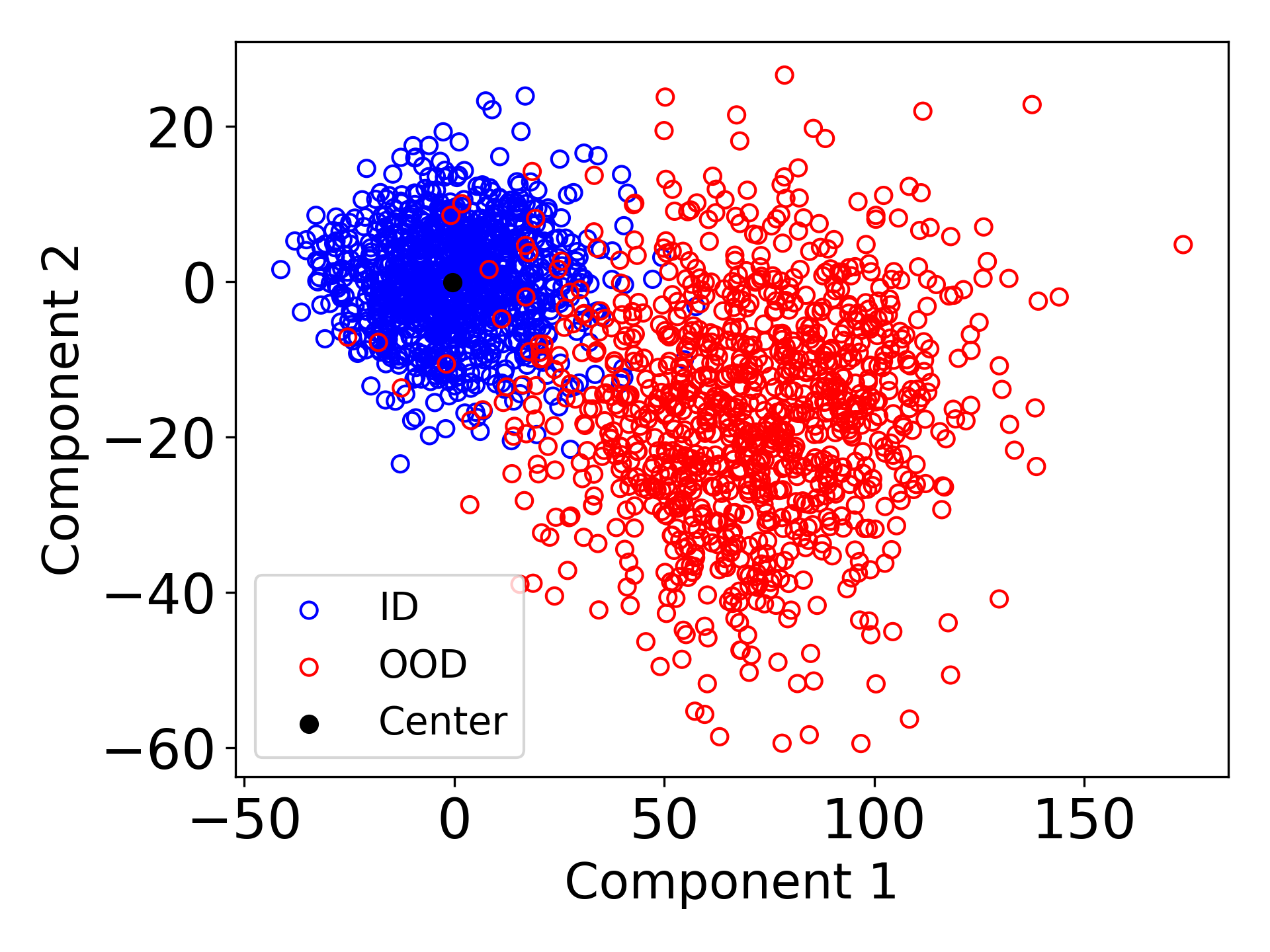}
        \caption{}
    \end{subfigure}}
    \caption{Effect of the proposed Feature Pooling strategy on differentiating between ID and OOD (LSUN) in feature space of ResNet model, visualized in 2D using PCA. (a) Original feature space and (b) Pooled feature space obtained for ID dataset CIFAR-10, (c) Original feature space and (d) Pooled feature space obtained for ID dataset CIFAR-100. The black dots represent the center of the feature space. }
    \label{effect}

\end{figure*}

Let $\phi(\textbf{x})$ be the vector representation for an input, $\textbf{x}$ obtained from layer $\ell$, constructed by applying two pooling operators on the extracted features maps, \emph{indexed-pooling} and \emph{max-pooling}. Using the above two pooling operators $\Gamma_1$ and $\Gamma_2$, therefore, the construction of $\phi(\textbf{x})$, for an input $\textbf{x}$, can be written as:
\begin{eqnarray}
    \phi(\textbf{x}; \ell, \gamma) &= (\Gamma_2 ~\circ ~ \Gamma_1) (\textbf{x}) 
\end{eqnarray}
\noindent 
where $\circ$ means $(f \circ g)(x) = f(g(x))$ [composite function].  We show in Figure~\ref{effect}, the segregation between ID and OOD samples obtained in feature space after executing the above feature selection operations.
%And, $\Gamma_2$ corresponds to \emph{max-pooling} extracted feature maps with a sliding window of size $k \times k$ with a stride of length $k$ in both axes.

\textbf{Feature Aggregation: }
 We compute $\phi(\textbf{x})$ for input all inputs $\textbf{x} \in \textbf D_{in}$ to represent the entire ID space as $\phi(\textbf D_{in})$. We then find a set of aggregated information for the entire ID space. For that, we at first find the aggregated cluster \emph{centroid}, denoted by $\textbf{c}$, of the feature space. The centroid is defined as follows:
\begin{equation}
\textbf{c} = \text{MEAN}(\phi(\textbf{x})) \text{ for } \textbf{x} \in \textbf D_{in}
\end{equation}
%After obtaining the combined feature space $\phi(\textbf x)$ for $\textbf x \in D_{in}$, we find the aggregated information from for the entire ID space. For that, we at first find the aggregated cluster center, denoted by $\textbf{c}$, of the feature space. The cluster center is defined as follows.

\noindent where {MEAN} computes the element-wise mean of the the collection of vectors obtained from $\textbf D_{in}$. 
The centroid, $\textbf{c}$, ultimately designates a center position of the ID space around which all ID samples position themselves in a close proximity. In that, the distance between the center and $\phi(\textbf{x})$ for any ID sample $\textbf{x}$ should follow a low-variance distribution. This distance is denoted as $\kappa(\textbf{x})$, which is defined as the Euclidean distance between $\phi(\textbf{x})$ and $\textbf{c}$:
\begin{equation}
    \kappa(\textbf{x}) = \| \phi(\textbf{x}) - \mathbf{c}\|
%      \kappa(\textbf{x}) = \left(\phi(\textbf{x}) - \textbf{c}\right)^T \mathbf{\Sigma}^{-1}\left(\phi(\textbf{x}) - \textbf{c}\right) 
\end{equation}

%\noindent for some co-variance matrix $\mathbf{\Sigma}$ defined on the features space. Note that if $\mathbf{\Sigma}^{-1}$ is set an identity matrix, the distance function becomes a regular Euclidean distance or L2-norm.

We hypothesize that since the centroid is a pre-determined value calculated using features of ID samples, distance, $\kappa(\textbf{x})$, will take smaller value for any an ID sample than the distance obtained for an OOD sample. We observe from Figure~\ref{effect} that feature space $\phi (\textbf x)$, when visualized in 2D, validates our hypothesis. That requires us to find a suitable a threshold value on this distance value based on which ID and OOD sample can be separated out. This is what we do next. 

 %We now define distance between an input sample, $\textbf{x}$ and the aggregated cluster center of the feature space, $\textbf{c}$, as follows.

%We also find another aggregated information out of $\phi(D_{in})$, the threshold on distance of $\phi(\textbf x)$ found for each input $\textbf x$ from centroid $\textbf c$, that gives a desired prediction confidence, $p$ on the ID samples. We hypothesize that since the centroid is a pre-determined value calculated using features of ID samples, distance from this centroid will be small for any other ID sample than the distance obtained for OOD samples. We recall that the distance for each sample $\textbf x$ is denoted by $\kappa (\textbf x)$, which is defined in equation-\ref{kappa}.

\iffalse
\begin{equation*}
    \kappa(\textbf{x}) = \left(\phi(\textbf{x}) - \textbf{c}\right)^T \mathbf{\Sigma}^{-1}\left(\phi(\textbf{x}) - \textbf{c}\right)   
 % \kappa(\textbf{x}) = \left(\phi(\textbf{x}) - \textbf{c}\right)^T \left(\phi(\textbf{x}) - \textbf{c}\right) 
  %\kappa(\textbf{x}) =\| \left(\phi(\textbf{x}) - \textbf{c}\right)\|
  %^T \left(\p
\end{equation*}
%\textcolor{red}{MENTION kappa(x) again}
\fi
We empirically find a threshold value, $\Theta$, that detects ID samples with some confidence $p$, such that $p$ fraction of the ID training samples have $\kappa(\textbf{x}) \leq \Theta$. That is:
\begin{equation}
    p = \frac{\sum_{\textbf{x} \in \textbf D_{in}} [\kappa(\textbf{x}) \leq \Theta]}{|\textbf D_{in}|}
\end{equation}

%\begin{eqnarray}
%  \Theta &= \inf\{p \leq F_\kappa(\kappa (\mathbf x))\}
%\end{eqnarray}
%where $F_{\kappa}(.)$ represents the Cumulative Distribution Function (CDF) of $\kappa(\textbf x)$ and $\inf$ represents the infimum function that returns the minimum value of $\kappa (\mathbf x)$ from the set of values whose CDF exceeds $p$.
We usually set $p = 0.95$. This is actually the expected TPR (True Positive Rate) of the OOD detector that we expect (the detector's capability to detect a true ID sample as ID). We note that, tuning the value of this hyperparameter $\Theta$ does not require exposure to any a priori known OOD samples.

\textbf{OOD Detection during Inference: } During inference, for an incoming input sample $\textbf x$, we first pass the input into the model up to layer $\ell$ and extract the intermediate output from that layer. We then choose $\mathcal N$ best maps from that intermediate output (specified by the binary index-vector $\gamma$). Then we do max-pooling on that space to find $\phi (\textbf x)$. After that we find the distance of $\phi (\textbf x)$ from the centroid, $\textbf{c}$, as $\kappa (\textbf x)$ and compare this value with the predefined threshold $\Theta$ for detecting if the sample is ID or OOD. 
Let $\mathcal D(\textbf{x})$ denote the detector output for input $\textbf{x}$, which can be obtained as:
%We denote the combined detector by $\mathcal D$. then inference on sample $\textbf x$ using $\mathcal D$ can be written as:
\begin{equation}
%\label{eq1}
   \mathcal D(\textbf{x}) = \begin{cases}
  1, & \text{if } \kappa(\textbf{x}) \leq \Theta, \\
  0, & \text{otherwise}.
\end{cases}
\end{equation}
So the inference is very fast and since our chosen layer $\ell$ is very shallow (unlike~\cite{hendrycks2016baseline, liang2018enhancing, lee2018simple} that detect OOD samples at the last layer), we can reject the extraneous OOD samples, way before lots of unnecessary computations are done on the sample, which would lead nowhere. Hence our ID detection approach gives higher throughput during batch inference. Besides we do not retrain the classifier model unlike~\cite{hsu2020generalized}. Also, we detect using features collected from a single layer only unlike~\cite{yu2020convolutional, lee2018simple}, without preprocessing input samples unlike~\cite{hsu2020generalized, liang2018enhancing, lee2018simple} and without using OOD samples for validation unlike~\cite{lee2018simple,liang2018enhancing}. The comparison of EARLIN with other approaches is summarized in Table~\ref{tab:comparison}.
%Compared to the other popular ID detection approaches, our approach called EARLIN adds merits in that its both lightweight in terms if both training and inference, fast and unbiased.
%We don't retrain the original deep learning classifier for ID detection. Also, unlike the other most popular approaches for ID detection~\cite{hendrycks2016baseline, liang2018enhancing, lee2018simple}, that detect ID samples at the last layer of the network, we detect ID samples very fast into the network pipeline, by using intermediate outputs from the very shallow layers. Besides, we detect using outputs from one layer only, unlike~\cite{yu2020convolutional, lee2018simple}. Additionally, we do not preprocess input images by adding loss-based perturbation, unlike~\cite{liang2018enhancing, lee2018simple}. Hence our approach gains speedup from this also. Moreover, the training phase involved with EARLIN is very minimal. It is just finding some parameters initially using a subset of samples on which the classifier model was trained. We do not expose any ID sample to the framework prior to testing the approach, whereas both \cite{liang2018enhancing} and \cite{lee2018simple} expose test ID for fine-tuning the values of the hyperparameters, rendering the resulting detector biased towards those samples.

%section
\section{Collaborative Inference based on EARLIN}
\label{collinf}
\begin{wrapfigure}{r}{0.65\textwidth}
%\begin{figure}

\centering

    \fbox{\includegraphics[width=1\linewidth]{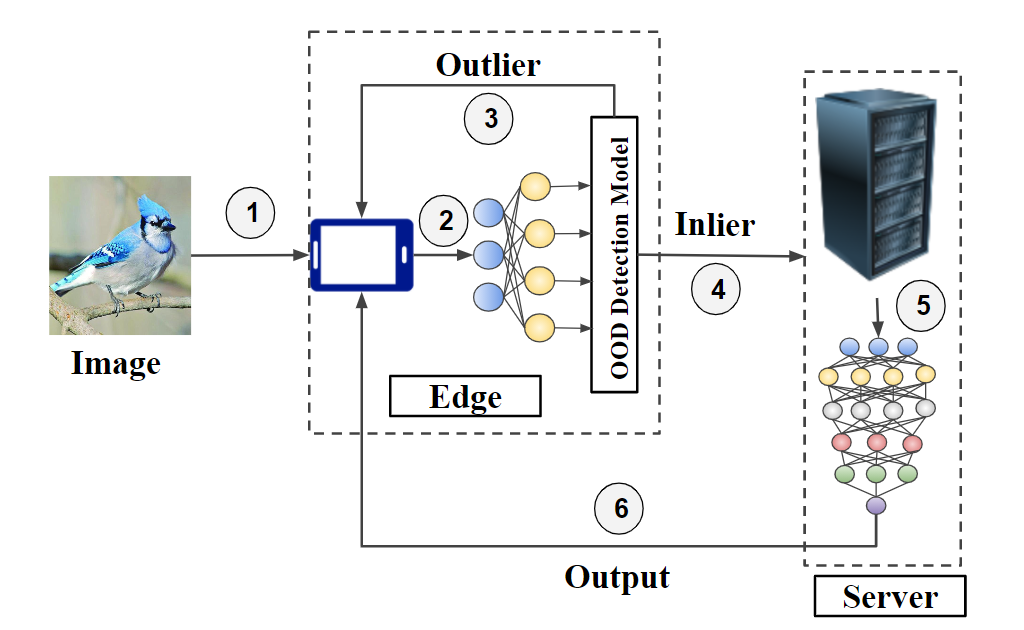}}
    \caption{Collaborative Inference Scheme.}
    \label{fig:coll}

%\end{figure}
\end{wrapfigure}
Based on our proposed OOD detection technique, we develop a setup for collaborative inference as a collaboration between an edge device and a server (this server can be in the cloud or can be a nearby edge resources, such as Cloudlet~\cite{verbelen2012cloudlets}, Mobile Edge Cloud (MEC)~\cite{liu2017mobile}, we generically refer to it as ``server''). Deep learning models usually have large memory and storage requirements, and hence are difficult to deploy in the constrained environment of the edge devices. Thus, edge devices make remote call to the server devices for inference. If the incoming image is Out-of-Distribution, making such call is useless since the model would not be able to classify the image. Hence, we both save resources and make more precise recognition by not allowing to call when input image is OOD. Since our detection model, consisting of the first few layers of the network architecture, is very lightweight, we deploy the detection pipeline in the edge device. Then, if the image is detected as ID, we send the image to the server for classification. Otherwise, we report that the image is OOD and hence not classifiable by the model. We thus save resource by not sending the OOD images to the server. The schematic diagram of the framework appears in Figure~\ref{fig:coll}. We note that we send the original image, instead of the intermediate layer output to the server, when the sample is detected as ID. This is because intermediate layer outputs from deep learning models at the shallower layers are often significantly higher in dimension than the original images. With that, we save considerable upload bandwidth. As the servers are usually high-end machines, repeating the same computation up to layer $\ell$ adds very nominal overhead compared to the volume of data to be uploaded. Moreover, $\ell$ is a very shallow layer (below 10\% from the input layer) as reported in Table~\ref{tab:models}. We also note that all model parameters are estimated/trained in the cloud using base model and the training datasets, and the resultant detector model is deployed on the edge device.

\textbf{Overall Accuracy of OOD Detector and Deep Learning Classifier: }
%\label{def1}
Traditionally the performance of deep learning classification models are reported in terms of accuracy to establish how well they perform. \textit{Accuracy} is defined as the ratio or percentage of samples classified correctly by the model, \emph{given that} every sample comes from In-Distribution (ID). Let us denote $\textbf{x} \in ID$ to indicate if an input $\textbf{x}$ \emph{truly} belongs to ID and $\textbf{x} \in OOD$ to denote if $\textbf{x}$ truly belongs to OOD (ideally, $\textbf{x} \in OOD$ is logically equivalent to $\textbf{x} \notin ID$). In terms of probability expression, the  $Accuracy$ (written as \emph{acc} in short) of a model, $\mathcal M$, can be written as:
\begin{equation*}
   acc_{\mathcal M} \delequal \mathcal P(\mathcal M(\textbf x) = y(\textbf{x}) \mid \textbf{x}\in  {ID})
\end{equation*}
where $\mathcal M(\textbf{x})$ represents the classification output of model $\mathcal M$ and $y(\textbf{x})$ represents the true class label of input $\textbf{x}$, $\mathcal P(E)$ denotes probability of event $E$. On the other hand, the performance of an OOD detector can be expressed in term of two metrics: True Positive Rate (TPR) and True Negative rate (TNR). TPR is the ratio of ID samples correctly classified as ID by the detector where TNR is the ratio of the true OOD samples detected as OODs. Let $\mathcal D$ denote a detector and $\mathcal D(\textbf{x})$ denote a binary output of the detector to indicate that whether input $\textbf{x}$ is detected as ID or OOD ($\mathcal D(\textbf{x}) = 1$ if detected as ID else 0). Consequently, the TPR and TNR values of a detector, $\mathcal D$, can be expressed as:
\begin{eqnarray}
   {TPR}_{\mathcal D} & \delequal & \mathcal P( \mathcal D (\textbf{x})= 1 \mid \textbf{x}\in ID) \\
   {TNR}_{\mathcal D} & \delequal & \mathcal P( \mathcal D (\textbf{x})= 0 \mid \textbf{x}\in OOD)
\end{eqnarray}

While the capability of the classification model ($\mathcal M$) and the OOD detector model ($\mathcal D$) can be expressed individually in terms of their respective performance metrics (that is, model classification accuracy, TPR and TNR), it is interesting to note how these three terms play a role in measuring the accuracy of the model and the detector combined. We refer to  this as the joint accuracy or \textit{overall accuracy}. We define the \textbf{Overall Accuracy} as the success rate of assigning \emph{correct} class labels to test inputs. That is, for an ID sample, this corresponds to assigning correct the class label to the input whereas, for an OOD sample, this corresponds to detecting it as an OOD (OOD samples do not have any correct class label other than being flagged as OOD). Let us use $\mathcal M \oplus \mathcal D$ to denote the classification model and detector combined and we are interested to determine the accuracy of $\mathcal M \oplus \mathcal D$ as a function of its constituents. We observe that in addition to the above three metrics, the overall accuracy of the model and OOD detector combined is dependent on what fraction of inputs are actually OOD as opposed to ID as inputs are passed to the model. Let this ratio be denoted as $\rho$. Formally, $$\rho = \mathcal P(\textbf{x} \in OOD) \text{ and } 1-\rho = \mathcal P(\textbf{x} \in ID)$$

More specifically, given the accuracy of model $\mathcal M$ and the TPR and TNR values of the associated OOD detector, $\mathcal D$, the overall accuracy of $\mathcal M \oplus \mathcal D$ is given by:
\begin{equation}
%\begin{split}
    acc_{\mathcal M \oplus \mathcal D} =  acc_{\mathcal M} \times TPR_{\mathcal D} \times (1-\rho)
    +  TNR_{\mathcal D} \times  \rho \label{eq:acc_M+D}
%\end{split}
\end{equation}
%\end{definition}
\noindent
\iffalse
\textbf{Proof.}
The statement can be prove directly by observing the accuracy definition and by appropriate conditioning on the probability statements. An input, $\textbf{x}$, is accurately classified by $\mathcal M \oplus \mathcal D$ only if either of the two mutually exclusive conditions holds true: (a) $\textbf{x}$ is a truly an ID, it's detected as ID by the detector and the correct class label is assigned, (b) $\textbf{x}$ is truly an OOD and the detector detects it as OOD. That means, the ${Overall Accuracy_{\mathcal M \oplus \mathcal D}}$ is given by:
\begin{equation*}
\begin{split}
    acc_{\mathcal M \oplus \mathcal D} &= \mathcal P(\mathcal M(\textbf{x}) = y(\textbf{x}) \wedge \mathcal D(\textbf{x}) = 1 \wedge \textbf{x} \in ID) \\
    &+  \mathcal P(D(\textbf{x}) = 0 \wedge \textbf{x} \in OOD) \\
    &= \mathcal P(\mathcal M(\textbf{x}) = y(\textbf{x}) \mid \textbf{x} \in ID) \mathcal P(\mathcal D(\textbf{x}) = 1 \mid \textbf{x} \in ID) \mathcal P(\textbf{x} \in ID) 
    \\
    &+\mathcal P(D(\textbf{x}) = 0 \mid \textbf{x} \in OOD) \mathcal P(\textbf{x} \in OOD) 
    \\
    &= acc_{\mathcal M} \times TPR_{\mathcal D} \times (1-\rho) + TNR_{\mathcal D} \times \rho 
\end{split}
\end{equation*}

This ends the proof. \qed
\fi

The proof of the above equation is based on the fact that a correct output occurs when either of the two mutually exclusive events happen with respect to an input sample: (a) the input sample truly belongs to ID, and the detector also detects it as ID and the model correctly classifies it, (b) the input sample belongs to OOD and the detector detects this as OOD (detail appears in the supplementary document).  

%given in the supplementary section. This metric captures the performance of the setup, which is a combination of both model and the OOD detector. This characterization of the combined setup is one of the contributions of this paper. We note that this metric is actually the performance measure of the setup from the perspective of an end user. 
%The user does not care about how accurately model detects IDs, unless in conjunction with the model accuracy as well. To the end user, the ultimate class label, along with the tag for OOD, is visible. So having high $TPR$ but low $Accuracy$ and vice-versa cannot be the characteristic of a well-performing from the perspective of end user. Instead, the combined setup of model and detector should  have high TPR, TNR, and $Accuracy$, which will jointly result in higher $Overall \, Accuracy$, indicating that the model is performing well. 
%\end{definition}

\textbf{Performance and Cost Characteristics of the Collaborative Setup: }

\label{sec: perf_char}
As per Eq~(\ref{eq:acc_M+D}), the overall accuracy depends on four quantities: accuracy of the original model, TPR and TNR of the detector, and $\rho$ (fraction of samples being OOD in the inference workload). Without any detector in place (when TPR becomes 1 and TNR is 0), the overall accuracy of the model $acc_{\mathcal M \oplus \mathcal D} = acc_{\mathcal M} \times (1- \rho)$, sharply declines with $\rho$. With the detector combined, the overall accuracy of the model, in fact, improves at a rate of $TNR-{acc}_{\mathcal M} \times TPR$ with respect to $\rho$ (actually, the accuracy grows only when the slope is positive, that is, $TNR > {acc}_{\mathcal M} \times TPR$). In Section~\ref{sec:prot}, we demonstrate this.

%can be approximated as a linear function of $\rho$, assuming that $TPR$ is constant as per specification and $TNR$ is almost constant within a range for a particular setup of $\mathcal M$. Then the rate of change of $acc_{\mathcal M \oplus D}$ with respect to $\rho$ is the slope of the line and can be written as $TNR-{acc}_{\mathcal M} \times TPR$. 

\begin{figure}[!tbp]
  \centering
  \scalebox{1}{
  \begin{subfigure}[b]{0.46\linewidth}
    \includegraphics[width=\textwidth]{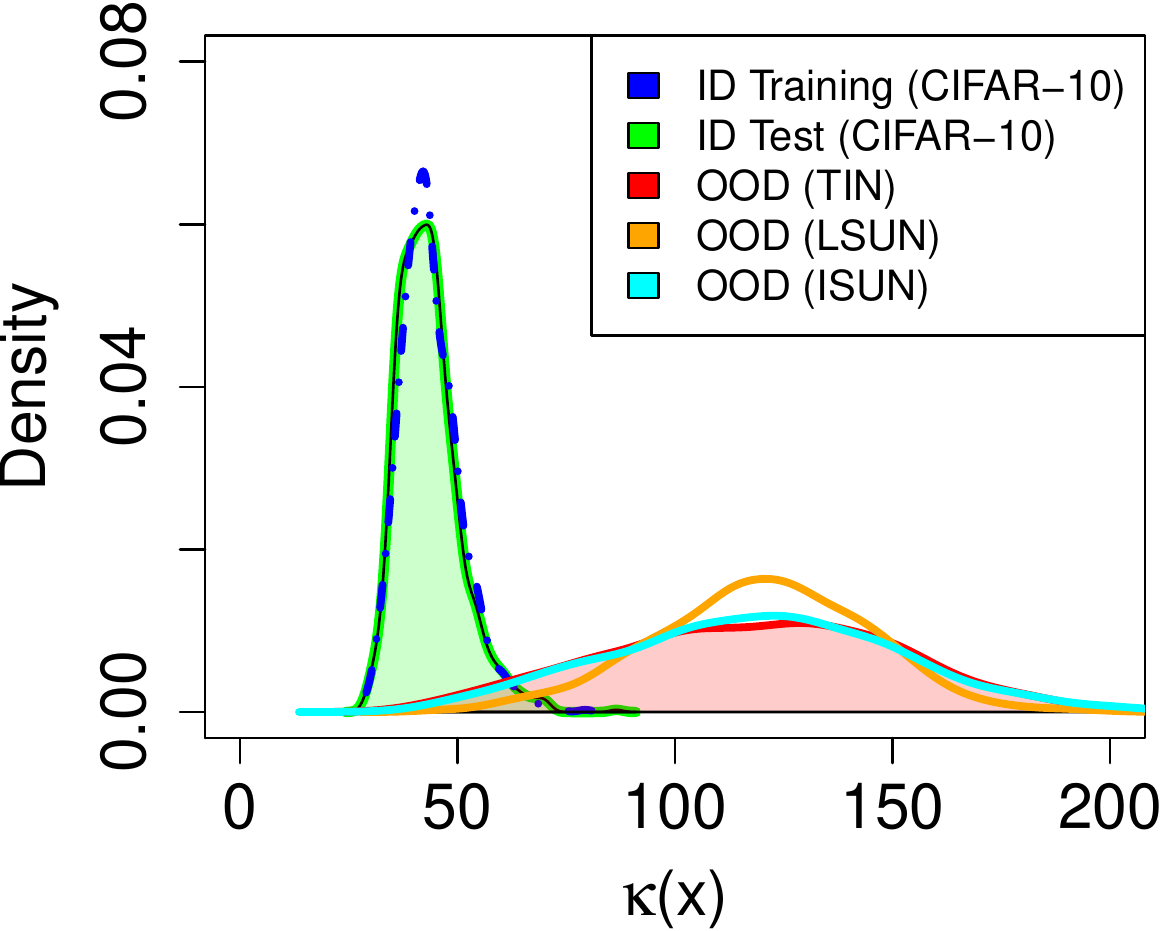}
    \caption{}
    \label{hist}
  \end{subfigure}
  %\hfill
  \begin{subfigure}[b]{0.49\linewidth}
    \includegraphics[width=\textwidth]{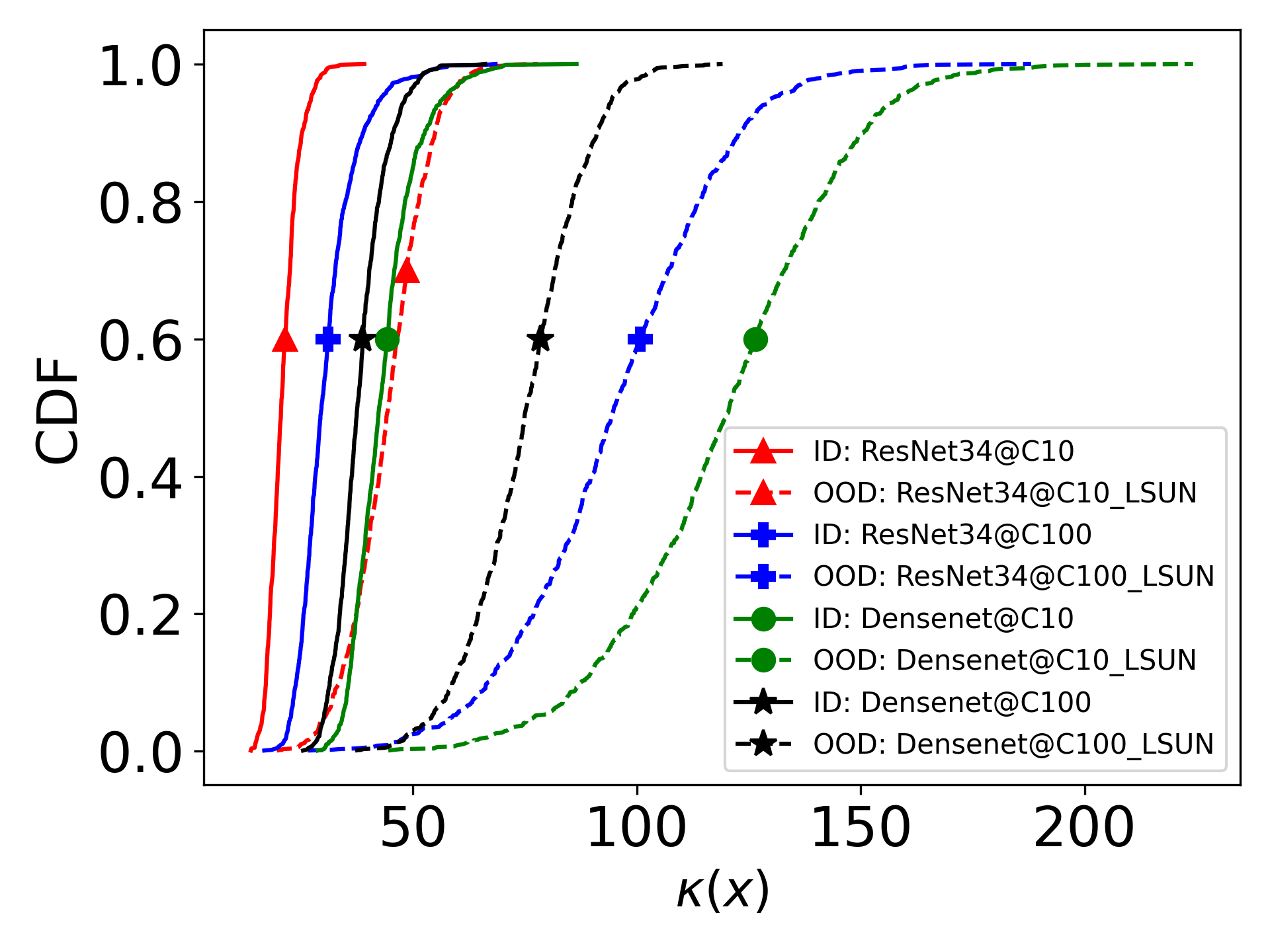}
    \caption{}
    \label{cdf}
  \end{subfigure}}
      \caption{(a) Histogram of $\kappa (\textbf x)$ that differentiates between ID and OOD samples (drawn from benchmark datasets) for Densenet pretrained on CIFAR-10. (b) CDF of $\kappa (\textbf x)$ that differentiates between ID and OOD samples with CIFAR-10 (C10) and CIFAR-100 (C100) as ID and the TinyImageNet (TIN), LSUN and iSUN dataset as OOD.} %for CIFAR-10 and CIFAR-100 with acronyms C10 and C100 respectively. TIN is acronym for TinyImagenet dataset. Notation $a@b\_c$ means CDF of $\kappa (\textbf x)$ of samples $\textbf x$ drawn from $c$ dataset on model $a$ pretrained on dataset $b$.}
      %and ResNet pretrained on CIFAR-100 datasets. TIN is acronym for TinyImagenet dataset.}
    \label{hist_CDF}
\end{figure}
%\subsection{Cost Characteristics of the Collaborative Setup}
%\label{sec: cost_char}
In EARLIN, as shown in Figure~\ref{fig:coll}, we send inputs to the server only when they are detected as ID by the lightweight OOD detector deployed at the edge. %, consisting of the first few shallow layers of model $\mathcal M$ deployed in the edge. 
Let $T_E$ be the time required for OOD detection at the edge, $T_C$ be the round-trip communication delay between the edge and the server, and $T_S$ be the time required for classifying the image at the server when sent. In that, when we encounter a sample that is detected as OOD (when $\mathcal D(\textbf{x})~=~0$), the time required is only $T_E$ (no communication to the server nor processing at the server). On the other hand, when an incoming sample is detected as ID (when $\mathcal D(\textbf{x})~=~1$), the inference latency becomes $T_E+T_C+T_S$. So, the time required for inference is closely associated with the ratio of OOD samples, $\rho$ and the precision with which the detector detects input samples as ID vs OOD. 
We can characterize the cost, in terms of latency, involved with each inference using $\mathcal M \oplus D$ as follows:
\iffalse
\begin{equation}
T_{\mathcal M \oplus D} = T_E + (T_C+T_S)\mathcal P(\mathcal D(\textbf x)=1)
\end{equation}
where, by conditioning on input sample being ID vs OOD, we obtain:
\begin{equation*}
\begin{split}
\mathcal P(\mathcal D(\textbf{x})=1) 
&= \mathcal P(\mathcal D(\textbf{x})=1 \mid \textbf{x} \in ID )\mathcal P(\textbf{x} \in ID) \\ 
&+ \mathcal P(\mathcal D(\textbf{x})=1 \mid \textbf{x} \in OOD )\mathcal P(\textbf{x} \in OOD)\\
 &= TPR_{\mathcal D} \times (1-\rho) + (1 - TNR_{\mathcal D})\times \rho  
\end{split}
\end{equation*}
which gives:
\fi
\begin{equation}
T_{\mathcal M \oplus D} = T_E + (T_C+T_S) \left(TPR_{\mathcal D} (1-\rho) + (1 - TNR_{\mathcal D}) \rho \right)  
\label{eq:latency}
\end{equation}

Similar to our performance indicator, $Overall \, Accuracy$, the cost characteristics of the setup, $T_{\mathcal M \oplus D}$, can also be approximated as a linear function of $\rho$ (OOD ratio). In general, the end-to-end inference latency declines as $\rho$ grows as OOD samples are intercepted by the OOD detector at edge thus reducing inference latency and saving communication resources. In particular, the inference latency declines at a rate of $(T_C+T_S) \times (FPR_{\mathcal D}-TPR_{\mathcal D})$, where $FPR = 1 - TNR$ with respect to $\rho$. More detailed performance characterization can be found in the supplementary section.

%section
\section{Experimental Evaluation of EARLIN}
\label{sec:exp}
In this section, we show how our proposed OOD detector, EARLIN, performs on standard pretrained models and benchmark datasets compared to the previously proposed approaches for OOD detection.

\iffalse
\begin{figure}[tbp]
\centering
\subfloat[subtitle of subfigure a]{\label{fig:a}\includegraphics[width=0.45\linewidth]{Images/CIFAR10_Densenet.jpg}}\qquad
\subfloat[subtitle of subfigure b]{\label{fig:b}\includegraphics[width=0.45\linewidth]{Images/CIFAR10_Resnet34.jpg}}\\
\subfloat[subtitle of subfigure c]{\label{fig:c}\includegraphics[width=0.45\textwidth]{Images/CIFAR100_Densenet.jpg}}\qquad%
\subfloat[subtitle of subfigure d]{\label{fig:d}\includegraphics[width=0.45\textwidth]{Images/CIFAR100_Resnet34.jpg}}%

\caption{
        TNR at 95\% TPR for different combinations of feature selection. (a) DenseNet and (b) ResNet34 pretrained on CIFAR-10, (c) DenseNet and (d) ResNet34 pretrained on CIFAR-100 }
        \label{fig:mean and std of nets}
\label{fig:effect}
\end{figure}
\fi
%\textbf{Choice of Hyperparameters}
%\label{hyper}

%In this work, we consider the confidence limit of 95\% on ID detection. We calculate the values for 95\% confidence threshold. 

\textbf{Evaluation Metrics of OOD Detection: } \textit{TNR and FPR at 95\% TPR: }
%This is the rate of detecting an OOD sample as OOD, which we define in Section-\ref{sec: perf_char}, when the TPR, also defined \ref{sec: perf_char}, is as high as 95\%.
This is the rate of detecting an OOD sample as OOD. Hence, $TNR = TN/(FP+TN)$, where FP is the number of OOD samples detected as ID and TN is the number of OOD samples detected as OOD. We report TNR values obtained when $TPR = TP/(FN+TP)$, TP being the number of ID samples detected as ID, is as high as 95\%. And FPR is defined as (1-TNR).
\begin{wraptable}{r}{0.50\textwidth}
%\begin{table}[!t]
\centering
\caption{Chosen Layer and Size of corresponding OOD detection models in pretrained Models}\label{tab:models}
%\scriptsize
%\scalebox{0.5}{
\adjustbox{width=\linewidth}{%
\begin{tabular}{c|c|c|c|c}
\toprule[2pt]
%\textbf{Size of Detector Model (in KB)}
\textbf{Model} & \# \textbf{of Layers} &\textbf{Chosen Layer}  &\multicolumn{1}{|p{2cm}|}{\centering Size of Detector \\ Model \\(in KB) } &\multicolumn{1}{|p{2cm}}{\centering Training \\Dataset}  \\
\midrule[2pt]
%VGG16 &16 &BN ($1st$) &175 &CIFAR-10  \\ \midrule[1pt]
%VGG16 &16 &BN ($1st$) &175 &CIFAR-100  \\ \midrule[1pt]
ResNet &34 &BN ($5^{th}$) &112 &CIFAR-10 \\ \midrule[1pt]
ResNet &34 &BN ($2^{nd}$) &55 &CIFAR-100 \\ \midrule[1pt]
%ResNet &44 &BN  ($4^{th}$) &89 &CIFAR-10 \\ \midrule[1pt]
%ResNet &44 &BN ($4^{th}$) &89  &CIFAR-100  \\ \midrule[1pt]
DenseNet &100 &BN  ($10^{th}$) &256 &CIFAR-10  \\ \midrule[1pt]
DenseNet &100 &BN ($10^{th}$) &256 &CIFAR-100  \\ 
\bottomrule[2pt]
\end{tabular}}%}
\end{wraptable}
%\textit{FPR at 95\% TPR: }
%This is the rate of detecting an OOD sample as OOD, which we define in Section-\ref{sec: perf_char}, when the TPR, also defined \ref{sec: perf_char}, is as high as 95\%.
%This is the rate of detecting an OOD sample as OOD. Hence, TNR = TN/(FP+TN), where FP is the number of OOD samples detected as ID and TN is the number of OOD samples detected as OOD. We report TNR values obtained when TPR = TP/(FN+TP), TP being the number of ID samples detected as ID, is as high as 95\%.
\textit{Detection Accuracy and Detection Error: }
This depicts the overall accuracy of detection and is calculated using formula $0.5 \times (TPR + TNR)$, assuming that both ID and OOD samples are equally likely to be encountered by the classifier during inference. And Detection Error is (1-Detection Accuracy). \newline
\textit{AUROC:} This evaluates area under the ROC curve.
\iffalse
\begin{figure}

    \centering
    \includegraphics[width=0.45\linewidth]{Images/Final_all_acc_comp_v7.png}
    \caption{Change in Performance of Collaborative Setup with ratio of OOD samples using models pretrained on CIFAR-100 dataset and TinyImagenet as OOD dataset.}
    
    \label{fig:accuracy_comparison}
\end{figure}
\fi
%\subsection{Results: EARLIN}
%\label{results}

% \usepackage{multirow}

\textbf{Results: }
We conduct experiments on Densenet with 100 layers (growth rate = 12) and ResNet with 34 layers pretrained on CIFAR-10 and CIFAR-100 datasets. Each of the ID datasets contains $50,000$ training images and $10,000$ test images. Summary of the pretrained models used in terms of their total number of layers, chosen layer $\ell$ for OOD detection, size of detector model $\mathcal D$, ID dataset on which the model was trained and the classification accuracy of the corresponding model are shown in Table~\ref{tab:models}. \par

%Please add the following packages if necessary:

%If the table is too wide, replace \begin{table}[!htp]...\end{table} with
%\begin{adjustwidth}{-2.5 cm}{-2.5 cm}\centering\begin{threeparttable}[!htb]...\end{threeparttable}\end{adjustwidth}

\begin{table*}[!tpb]

\centering
\caption{OOD detection performance on different datasets and pretrained models. Here MLCM stands for MALCOM~\cite{yu2020convolutional}, BASE for Baseline~\cite{hendrycks2016baseline}, ODIN for ODIN~\cite{liang2018enhancing} and MAHA for Mahalanobis~\cite{lee2018simple}. \textbf{bold} indicates best result.}
\label{tab:performance}

\scalebox{1}{
\adjustbox{max width=\textwidth}{%

%\begin{tabular}{lrr|rrrr|rrrr|rrrrr}
\begin{tabular}{lcc|ccccc|ccccc|ccccc}
\toprule
ID Dataset &Model &OOD & & TNR at $95\%$ TPR && && &Detection Accuracy && && &AUROC & & & \\
\midrule[1pt]
\iffalse
& & &Softmax~\cite{hendrycks2016baseline} &ODIN~\cite{liang2018enhancing} &Mahalanobis~\cite{lee2018simple} &EARLIN &Softmax~\cite{hendrycks2016baseline} &ODIN~\cite{liang2018enhancing} &Mahalanobis~\cite{lee2018simple} &EARLIN \\\midrule[1pt]
\multirow{3}{*}{CIFAR-10} & &TinyImagenet &64.47 &67.04 &68.24 &\textbf{79.80} &79.74 &81.02 &81.63 &\textbf{87.40} \\
& VGG 16&LSUN &72.60 &83.86 &82.54 &\textbf{91.60} &83.81 &89.44 &88.78 &\textbf{93.30} \\
& &iSUN &71.34 &82.02 &80.72 &\textbf{83.30} &83.18 &88.52 &87.87 &\textbf{89.15} \\
\midrule[1pt]
\multirow{3}{*}{CIFAR-10} & &TinyImagenet &63.85 &80.12 &31.62 &\textbf{93.90} &79.44 &87.58 &63.34 &\textbf{94.45} \\
&Resnet44 &LSUN &72.17 &91.28 &36.47 &\textbf{99.30} &83.60 &93.15 &65.74 &\textbf{97.15} \\
& &iSUN &70.02 &90.05 &33.83 &\textbf{97.10} &82.55 &92.55 &64.42 &\textbf{96.05} \\
\midrule[1pt]
\multirow{3}{*}{CIFAR-100} & &TinyImagenet &36.44 &49.34 &42.30 &\textbf{77.30} &65.76 &72.19 &68.66 &\textbf{86.15} \\
& VGG 16&LSUN &38.27 &54.30 &46.71 &\textbf{89.60} &66.64 &74.68 &70.89 &\textbf{92.30} \\
& &iSUN &35.58 &50.79 &43.22 &\textbf{80.90} &65.29 &72.90 &69.11 &\textbf{87.95} \\
\midrule[1pt]
\multirow{3}{*}{CIFAR-100} & &TinyImagenet &36.61 &55.91 &20.30 &\textbf{93.70} &65.82 &75.48 &57.65 &\textbf{94.35} \\
&Resnet44 &LSUN &38.28 &68.75 &21.93 &\textbf{98.80} &66.64 &81.89 &58.47 &\textbf{96.90} \\
& &iSUN &38.33 &64.40 &14.01 &\textbf{96.20} &66.67 &79.70 &54.51 &\textbf{95.60} \\
\midrule[1pt]
\fi
& & &MLCM&BASE&ODIN&MAHA&EARLIN &MLCM&BASE &ODIN &MAHA &EARLIN &BASE
 &ODIN &MAHA
 &MLCM
 &EARLIN\\\midrule[1pt]

\multirow{3}{*}{CIFAR-10} & &TinyImagenet &95.50 &81.20 &87.59 &93.61 &\textbf{97.50} &95.33 &88.10 &92.34 &94.38 &\textbf{96.25} &94.10 & 97.69  &98.29 &99.06 &\textbf{99.14} \\

&Densenet &LSUN & 96.78 &85.40 &94.53 &96.21 &\textbf{99.30} &96.07 &90.20 &94.91 &95.78 &\textbf{97.15} &95.50 &98.85 &98.91 &99.23 &\textbf{99.85} \\

& &iSUN &95.59 &83.30 &91.81 &93.21 &\textbf{97.60} &95.41 &89.15 &93.82 &94.17 &\textbf{96.30} &94.80 &98.40 &97.98 &99.04 &\textbf{99.37}\\
\midrule[1pt]
\multirow{3}{*}{CIFAR-10} & &TinyImagenet &\textbf{98.10} &71.60 &70.39 &97.53 &93.92 &\textbf{96.92} &83.30 &85.80 &96.55 &94.46 &91.00 &91.88 &99.43 &\textbf{99.56} &97.54 \\
&Resnet34 &LSUN &\textbf{99.04} &71.70 &81.94 &98.83 &98.00 &\textbf{97.65} &83.35 &90.01 &97.58 &96.5 &91.10 &95.55 &99.64 &\textbf{99.70} &99.55\\
& &iSUN &\textbf{98.25} &71.90 &77.89 &97.64 &95.19 &\textbf{96.94} &83.45 &88.4 &96.66 &95.09 &91.00 &94.26 &99.47 &\textbf{99.59} &98.74 \\
\midrule[1pt]
\multirow{3}{*}{CIFAR-100} & &TinyImagenet &87.12 &47.90 &53.88 &80.37 &\textbf{92.60} &91.65 &61.45 &81.32 &88.40 &\textbf{93.80} &71.60 & 89.16 &93.64 &97.21 &\textbf{98.04} \\
&Densenet &LSUN &90.46 &49.70 &60.77 &85.74 &\textbf{98.10} &92.87 &62.35 &84.51 &90.85 &\textbf{96.55} &70.80 &92.06 &95.82 &97.61 &\textbf{99.96} \\
& &iSUN &88.29 &47.30 &54.85 &81.78 &\textbf{94.00} &92.04 &61.15 &82.51 &89.30 &\textbf{94.50} &69.60 &90.29 &94.81 &97.34 &\textbf{98.61} \\
\midrule[1pt]
\multirow{3}{*}{CIFAR-100} & &TinyImagenet &92.88 &31.00 &64.48 &91.76 &\textbf{95.40} &94.10 &58.00 &85.77 &93.56 &\textbf{95.20} &67.10 &93.06 &98.28 &98.54 &\textbf{98.55} \\
&Resnet34 &LSUN &94.76 &35.30 &64.95 &95.31 &\textbf{99.20} &94.92 &55.15 &86.09 &95.22 &\textbf{97.10} &65.60 &93.39 &98.81 &98.71 &\textbf{99.74} \\
& &iSUN &92.36 &36.70 &63.03 &91.98 &\textbf{97.10} &93.81 &55.85 &85.33 &93.76 &\textbf{96.05} &65.60 &92.76 &98.27 &98.24 &\textbf{99.22} \\
\bottomrule[1.2pt]
\end{tabular}
}}
\label{earlin_results}
\end{table*}

\begin{figure*}[!htpb]
        \centering
        \begin{subfigure}[b]{0.24\linewidth}
            \centering
            \includegraphics[width=\textwidth]{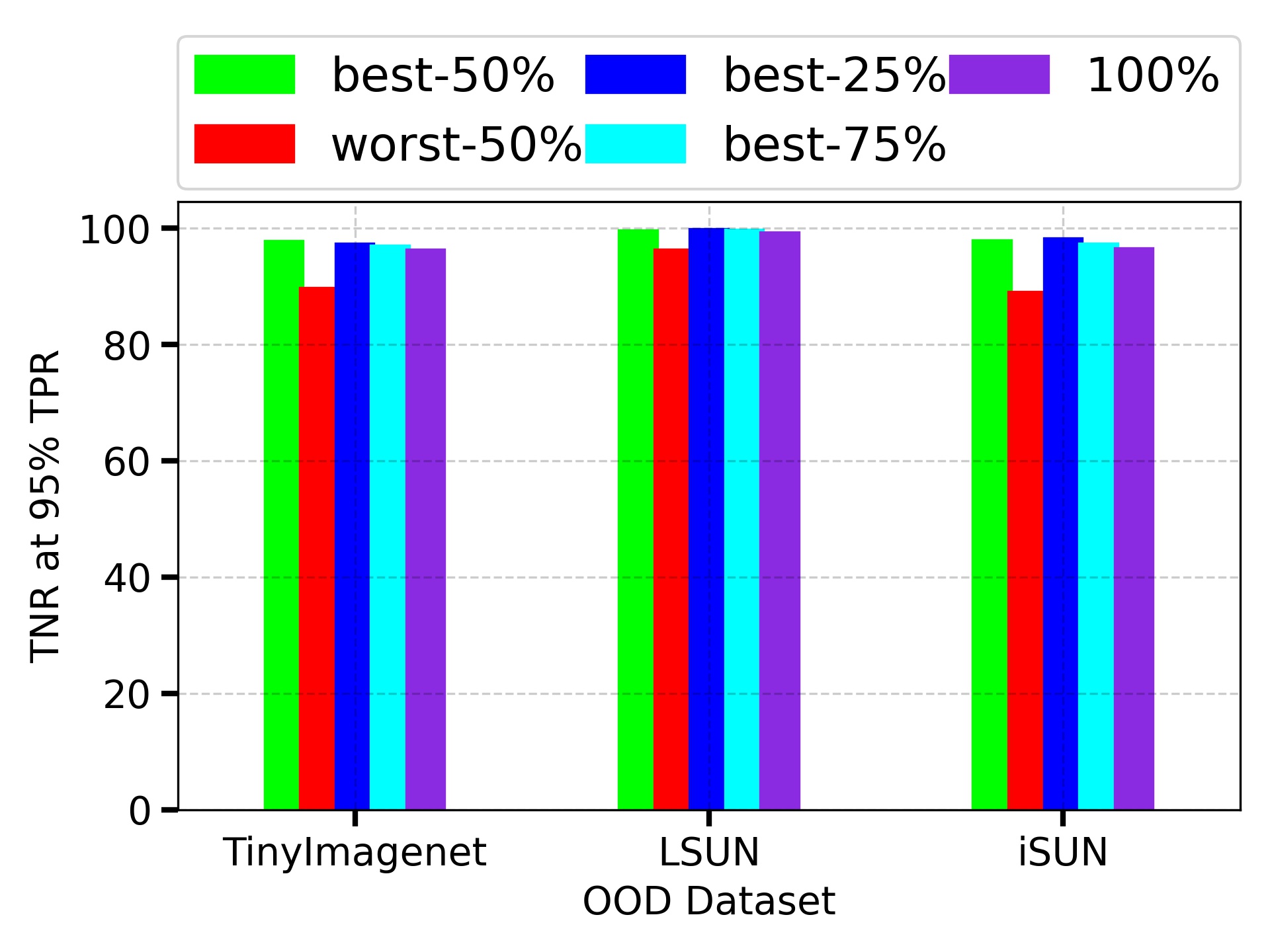}
            \caption[Network2]%
                
            \label{fig:mean and std of net14}
        \end{subfigure}
        %\hfill
        \begin{subfigure}[b]{0.24\linewidth}  
            \centering 
            \includegraphics[width=\textwidth]{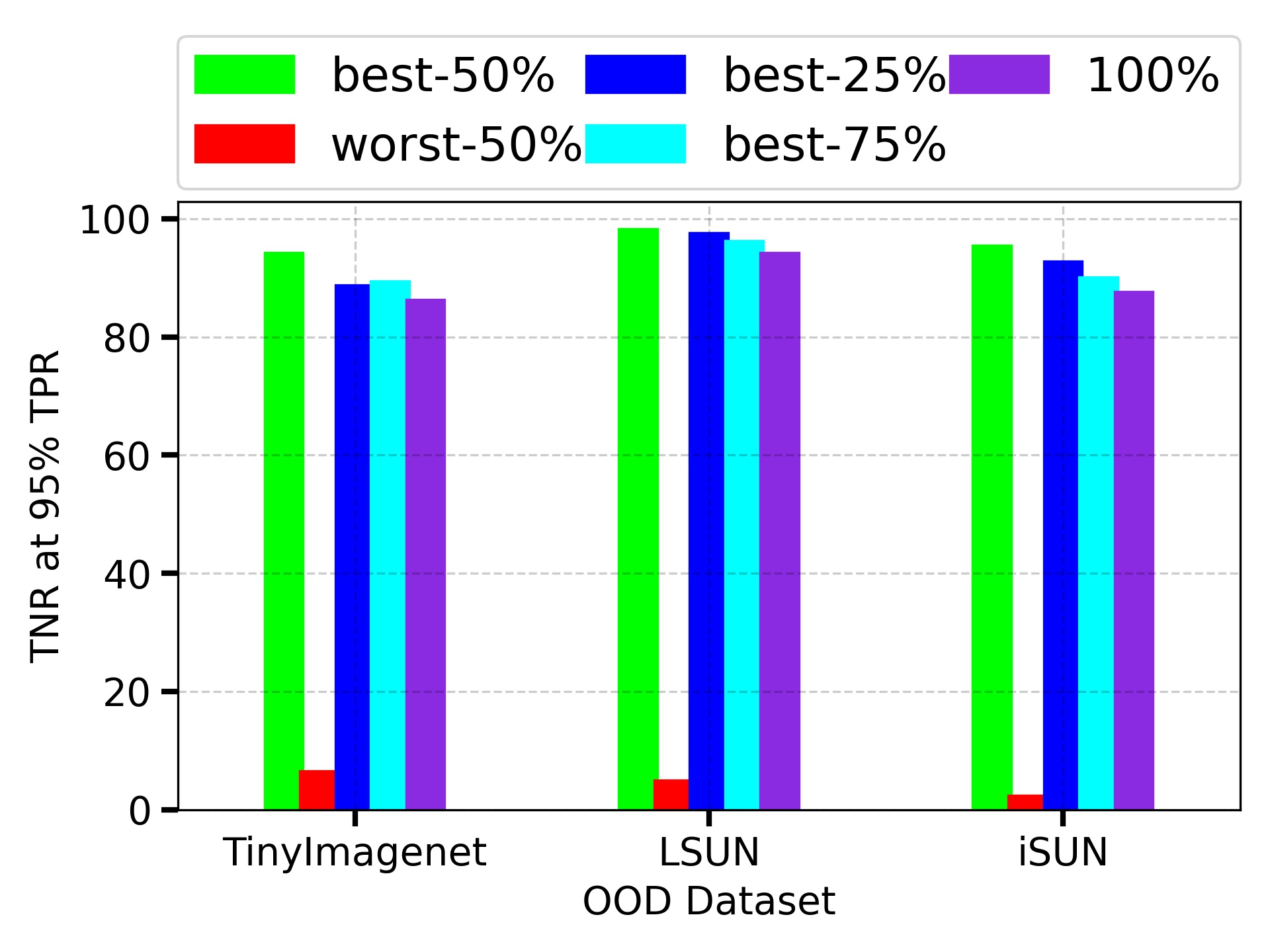}
            \caption[]%
                
            \label{fig:mean and std of net24}
        \end{subfigure}
        %\vskip\baselineskip
        \begin{subfigure}[b]{0.24\linewidth}   
            \centering 
            \includegraphics[width=\textwidth]{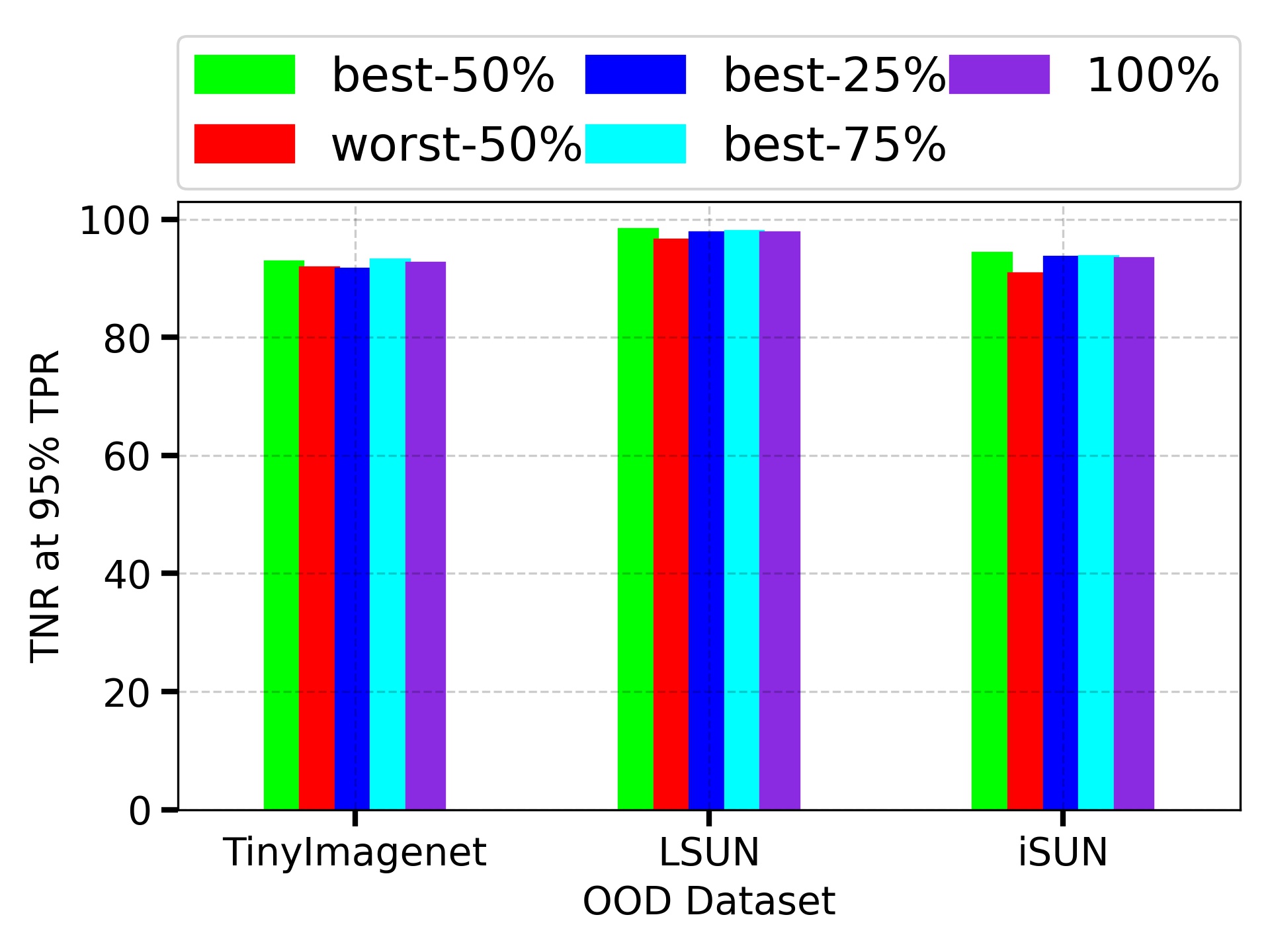}
            \caption[]%
              
            \label{fig:mean and std of net34}
        \end{subfigure}
        %\hfill
        \begin{subfigure}[b]{0.24\linewidth}   
            \centering 
            \includegraphics[width=\textwidth]{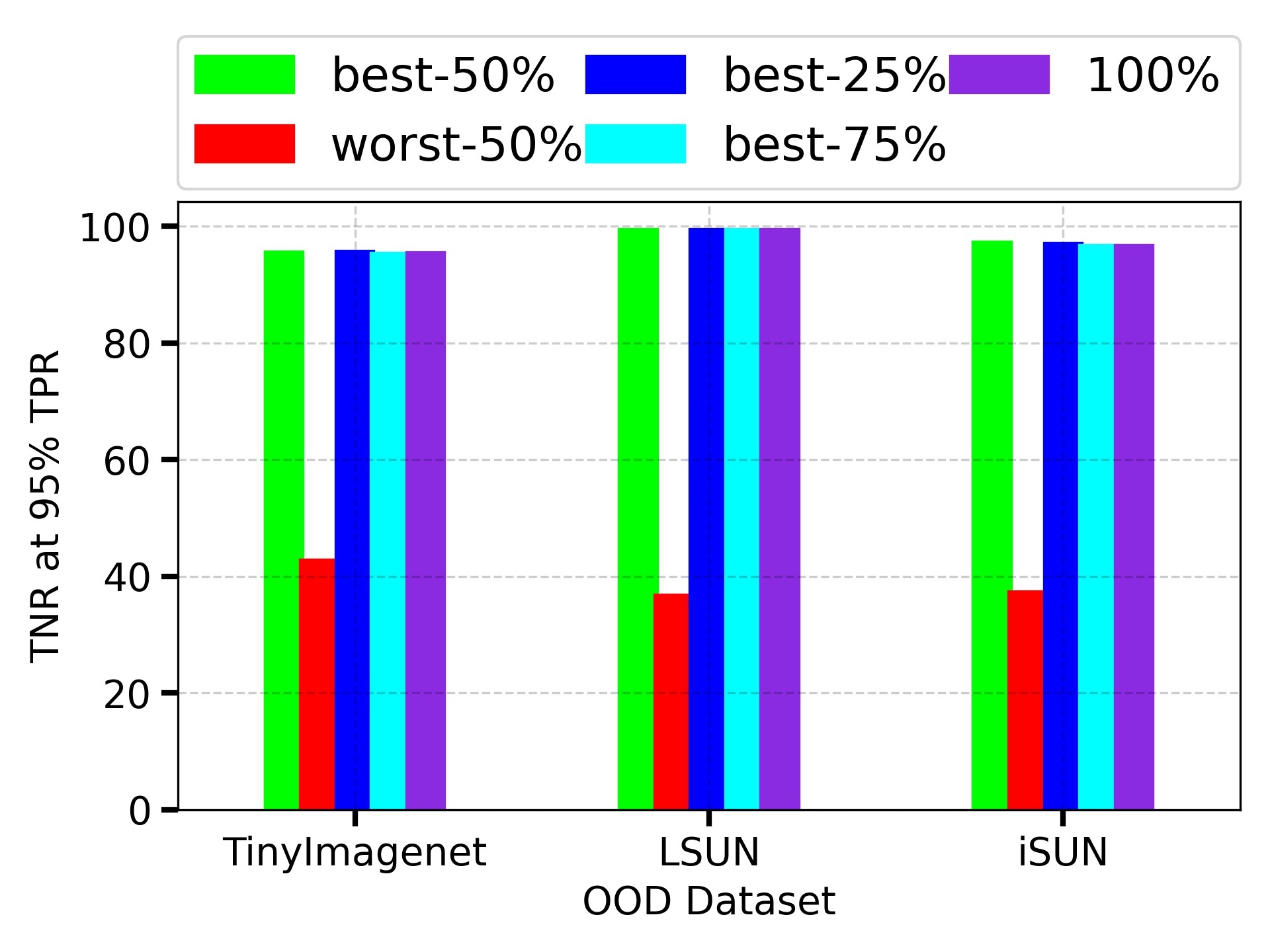}
            \caption[]%
              
            \label{fig:mean and std of net44}
        \end{subfigure}
        \caption{
        TNR at 95\% TPR for different combinations of feature selection. (a) DenseNet and (b) ResNet34 pretrained on CIFAR-10, (c) DenseNet and (d) ResNet34 pretrained on CIFAR-100 }
        \label{fig:mean and std of nets}
    \label{fig:effect}
\end{figure*}

In Table~\ref{earlin_results}, we show the TNR (at 95\% TPR) and Detection Accuracy of our approach. We compare our results with those obtained using previously proposed approaches, Baseline~\cite{hendrycks2016baseline}, ODIN~\cite{liang2018enhancing} Mahalanobis Detector~\cite{lee2018simple} and MALCOM~\cite{yu2020convolutional} on benchmark datasets~\cite{ooddata} TinyImagenet, LSUN and iSUN, popularly used for testing OOD detection techniques. It is to be noted that, we did not implement the earlier approaches (except Baseline), rather compare with the results reported in~\cite{yu2020convolutional} by using the same experimental setup. We see from the results in Table~\ref{earlin_results} that EARLIN performs better than the previous approaches in most of the cases. We report another set of results in Table-\ref{earlin_results2}, where we compare performance of EARLIN against DeConf~\cite{hsu2020generalized} on DenseNet model pretrained on CIFAR-10 and CIFAR-100 datasets, in terms of metrics TNR at 95\% TPR and AUROC. We note that we did not obtain results in the experimental setting on ResNet34 pretrained models in~\cite{hsu2020generalized}. We see from the results in Table~\ref{earlin_results2} also that EARLIN performs better than the previous approaches in most of the cases. We report yet another set of experimental results on VGG16 and ResNet44 models pretrained on CIFAR-10 and CIFAR-100 in the supplementary file.
\begin{wraptable}{r}{0.50\textwidth}
\centering
\caption{OOD detection performance of EARLIN compared to DeConf~\cite{hsu2020generalized}. \textbf{bold} indicates best result.}
%\label{tab:performance}
%\scriptsize
\scalebox{1}{
\adjustbox{max width=0.5\textwidth}{%
\begin{tabular}{lrr|cc|ccc}\toprule
ID Dataset &Model &OOD & &TNR at $95\%$ TPR  &AUROC & \\\midrule[1pt]

& &  &DeConf~\cite{hsu2020generalized} &EARLIN  &DeConf~\cite{hsu2020generalized} &EARLIN \\\midrule[1pt]
\multirow{3}{*}{CIFAR-10} & &TinyImagenet  &95.80 &\textbf{97.50}  &99.10 &\textbf{99.14} \\
& DenseNet&LSUN  &97.60 &\textbf{99.30}  &99.40 &\textbf{99.85} \\
& &iSUN  &97.50 &\textbf{97.60}  &\textbf{99.40} &99.37 \\
\midrule[1pt]
\multirow{3}{*}{CIFAR-100} & &TinyImagenet  &\textbf{93.30} &92.60  &\textbf{98.60} &98.04 \\
&DenseNet &LSUN  &93.80 &\textbf{98.10}  &98.70 &\textbf{99.96} \\
& &iSUN  &92.50 &\textbf{94.00}  &98.40 &\textbf{98.61} \\

%\midrule[1pt]
\bottomrule[1.2pt]
\end{tabular}}}
\label{earlin_results2}
\end{wraptable}
To demonstrate the clear separation of ID and OOD samples based on the estimated distance, $\kappa(\textbf{x})$, in Figure~\ref{hist_CDF}, we show the density and the corresponding CDF of $\kappa(\textbf{x})$ obtained from various test ID and test OOD datasets. We observe that ID and OOD samples have separable distribution based on $\kappa(\textbf{x})$.

%We also show the histogram of the predicted distance, $\kappa(\textbf{x})$, obtained for EARLIN that differentiates between ID and OOD samples, in Figure~\ref{hist} and the corresponding CDF plots in Figure~\ref{cdf}.

%\end{table}

\begin{figure}
    \centering
    \includegraphics[width=\textwidth]{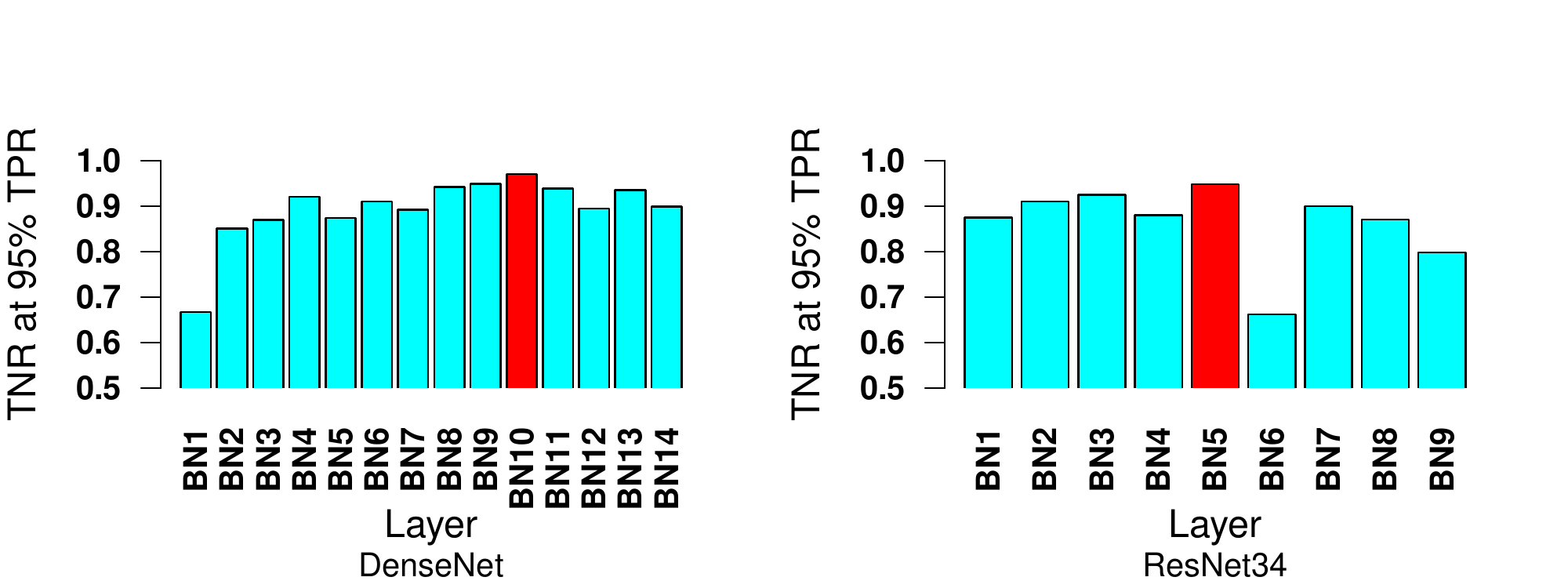}
    \caption{TNR at 95\% TPR for different choice of shallow BN layers in DenseNet and ResNet34 models pretrained on CIFAR-10. Random 1000 iSUN samples have been used as validation OOD data. Our chosen layer is shown in red in each case.}
    \label{fig:layer}
\end{figure}

\textbf{Ablation Studies: }
In order to detect samples as OOD as early as possible, we explore top (shallowest) 10\% layers of the pretrained models to find the separation between ID and OOD samples and report the layer $\ell$ that performed the best. It is to be noted that we consider only the Batch Normalization (BN) layers of the pretrained models as in these layers parameters sensitive to the ID dataset are learned during training. We show in Figure-\ref{fig:layer} how our end result, TNR at 95\% TPR varies for different choices of the shallow BN layers in ResNet34 and DenseNet models pretrained on CIFAR-10.
%We show the chosen layer $\ell$ for each pretrained model on which we experiment in Table-\ref{tab:models}.
%So far, we have chosen a suitable layer for obtaining the feature maps that give best separation between ID and OOD samples in a trial-and-error manner. for all the pretrained models on which we show the results, features are collected from selected layers that are among top $10\%$ of the total number of layers in the models.
 In Table~\ref{tab:models} we show our choice of layers for the models we considered. We observe that in all cases, the chosen layer is quite early in the network pipeline.

%so that we do not waste too many resources by doing computations on the OOD samples, that are beyond the classification capacity of the model. 
For finding the other hyperparameters, such as the number of maps $\mathcal N$, centroid $\textbf c$, and threshold $\Theta$, for each pretrained model, 20\% of the training ID samples were used as $\textbf D_{in}$ without using their corresponding classification labels. For each pretrained model, we set $\mathcal N$ to be half of the number of 2D feature maps at layer $\ell$. The threshold, $\Theta$, is set to 95\% ID detection confidence.
In Figure~\ref{fig:effect}, we show the effect of selecting different number of feature maps ($\cal N$), other than the default 50\% (half). 
%As mentioned earlier, the reported experimental results in Table~\ref{earlin_results} are for the case when 50\% of the 2D feature maps are selected based on $\psi$ values representing information content discussed in Section-\ref{infgain}. 
Figure~\ref{fig:effect} shows the TNR values at 95\% TPR for different datasets on different pretrained models, for different combinations of selecting 2D features: best 50\%, best 25\%, best 75\%, worst 50\% based on $\psi$ (Eq~(\ref{eq:psi})) and also all 100\%. We see that in almost all cases, selecting worst 50\% leads to the worst TNR for all datasets (more noticeable for ResNet34 models). Selecting top 50\% of the maps leads to either better or equivalent TNR, compared to selecting top 25\%, top 75\% and all 100\%. The choice of top 50\% of the 2D feature maps apparently produces the best results.

%section
\section{Prototype Implementation and Results}
\label{sec:prot}
%In this section, we discuss the experimental results obtained from the EARLIN-based Collaborative Inference Setup deployed in an edge-cloud distributed setting. 
%We show how the cost of inference and the overall accuracy change as we vary inference workload.
%, specially the mixture of ID and OOD samples. 

%We show how the cost and performance look like when tested on a sample population, with respect to both detection and latency involved with each sample. We also show how we improve in terms of both performance and cost. %Finally, we try a demo video application using our setup and report how the setup performs on real video data.

\begin{figure}[htpb]
     \centering
     \subfloat[]{
     \includegraphics[width=0.47\textwidth]{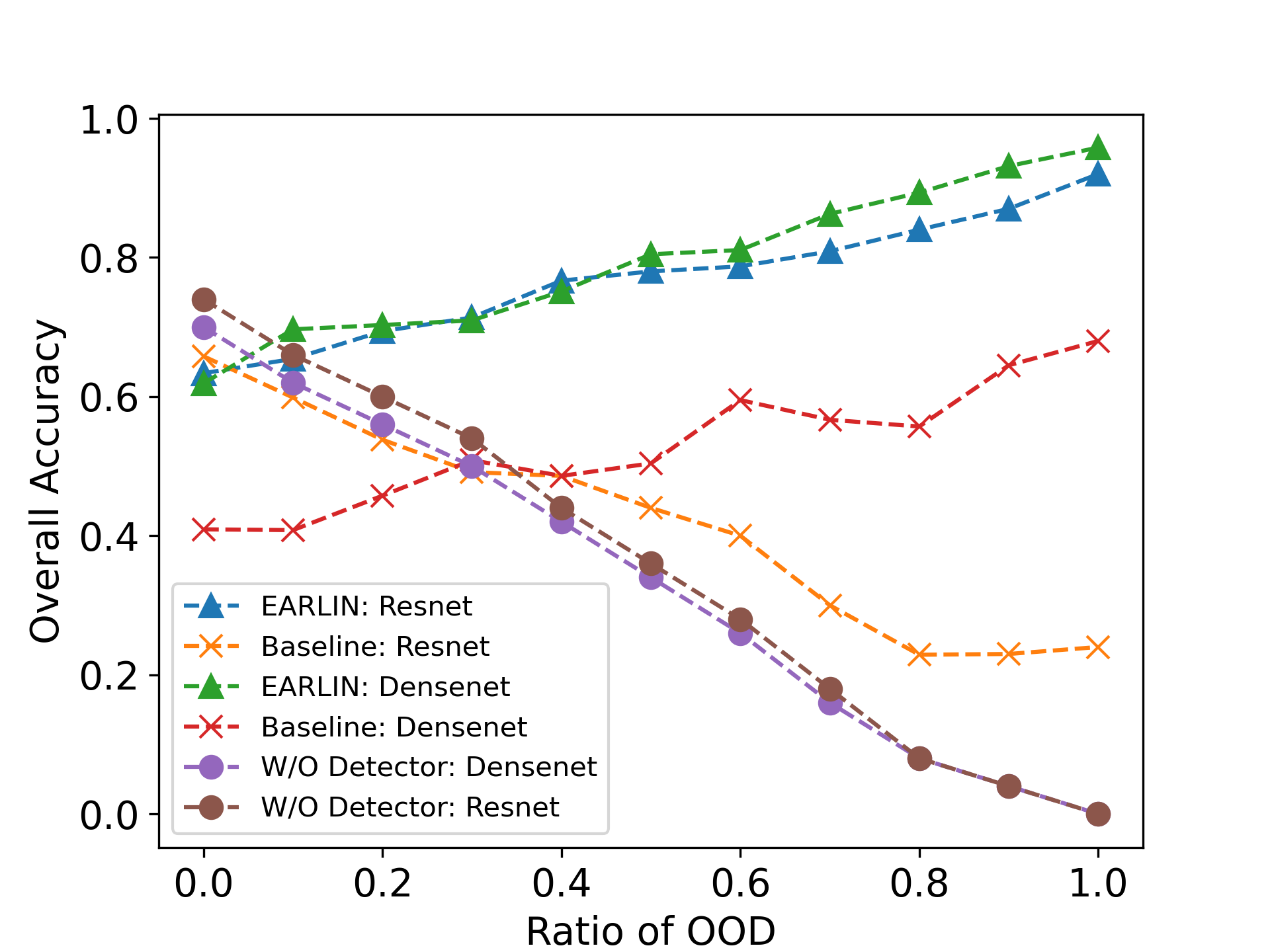}
     \label{acccomp}
     }\subfloat[]{
     \includegraphics[width=0.44\textwidth]{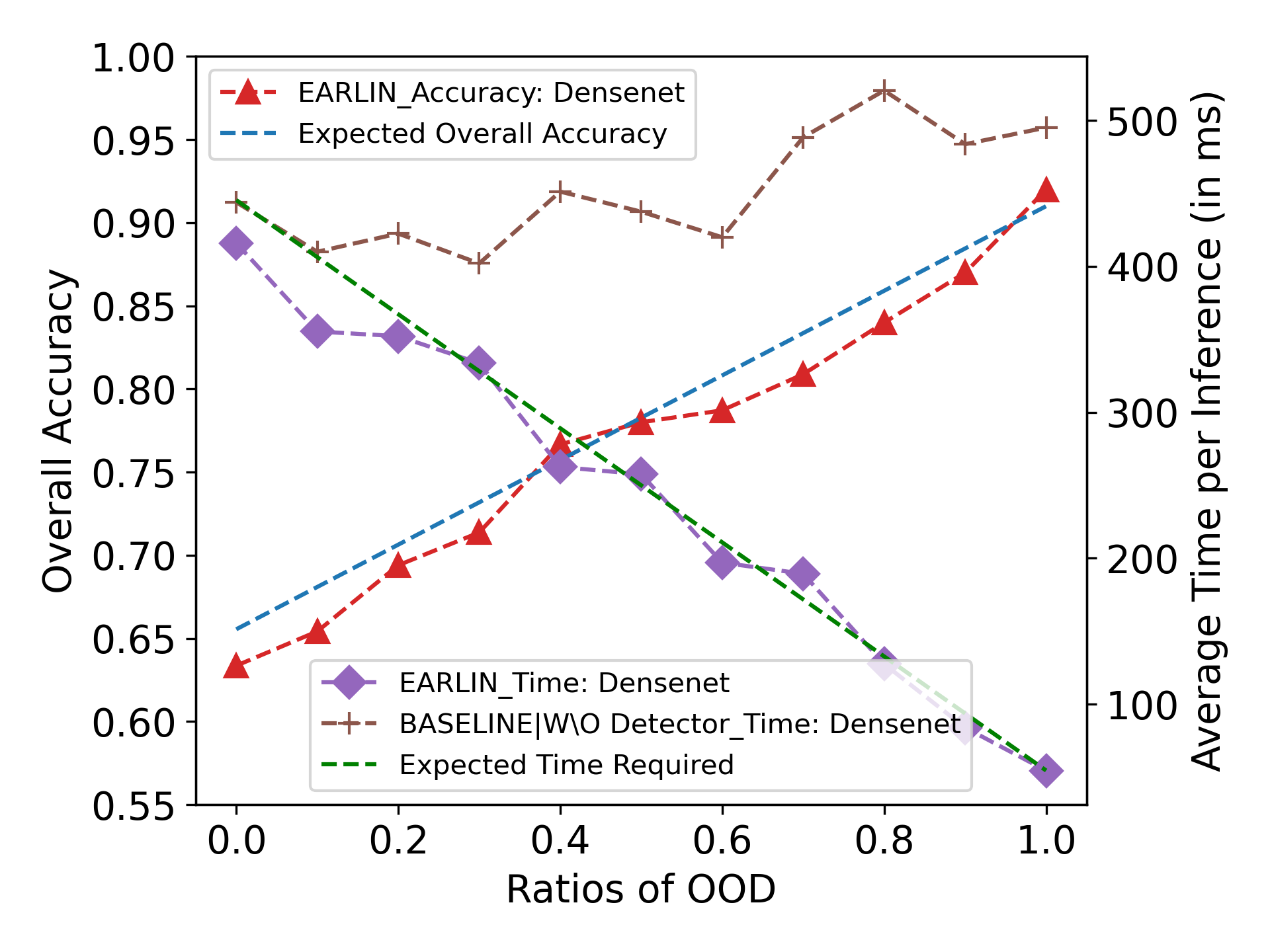}
     \label{latcomp}
     }
     \caption{Change in (a) Performance and (b) Cost vs Performance of Collaborative Setup with ratio of OOD samples using models pretrained on CIFAR-100 dataset and TinyImagenet as OOD dataset. }
    \label{acc_comp}
\end{figure}

\iffalse
\begin{figure}[!tpb]
    \centering
    \begin{subfigure}{0.45\textwidth}
        %\centering
        \includegraphics[width=\textwidth]{Images/Final_all_acc_comp_v7.png}
        
        %\caption{}
        \label{acccomp}
    \end{subfigure}%
    ~
    \begin{subfigure}{0.45\textwidth}
        %\centering
         \includegraphics[width=\textwidth]{Images/Final_acc_vs_time_v7.png}
        
        %\caption{}
        \label{latcomp}
    \end{subfigure}
     
    \caption{Change in (a) Performance and (b) Cost vs Performance of Collaborative Setup with ratio of OOD samples using models pretrained on CIFAR-100 dataset and TinyImagenet as OOD dataset. }
    \label{acc_comp}

\end{figure}
\fi
\textbf{Experimental Setup: }
We build a collaborative inference testbed where a client program with our EARLIN OOD detector runs on an edge device and the deep learning models are deployed on a server machine. Our client program runs on a desktop computer with a moderate CPU-only configuration (Intel®Core™ i7-9750H@2.60GHz CPU) and 32 GB RAM, a configuration similar to the edge setup described in~\cite{canel2019scaling}). The server program, developed using Flask and TensorFlow framework in Python, is deployed at the Google Cloud and is powered by Nvidia K80 GPU devices. For demonstrating the effectiveness of EARLIN, we deploy two CNN models in the cloud: (a) DenseNet with 100 layers and (b) ResNet with 34 layers (both are pretrained on CIFAR-100 with 70\% classification accuracy). We deploy their corresponding OOD detection part on the edge device. In all experiments, TinyImageNet dataset is used as OOD.
We set a threshold to have 95\% prediction confidence on ID samples, the condition we considered while reporting the results on EARLIN in Section-\ref{sec:exp}. Hence all the TPR, TNR, and \textit{Accuracy} (detector accuracy) values match those reported in Table~\ref{earlin_results}. We note that the mean latency for computations done at the edge ($T_E$) is $32.8\pm 15$ ms and at the server ($T_S$), it is $47.8\pm 25$ ms. The mean communication delay ($T_C$) is $186.5\pm52.12$ ms. We observe that latency at both edge and server is quite small. At the edge, we deploy a small portion of the model hence the latency is low. On the other hand, the server runs models on GPU resources so the inference time is small there. The communication delay to the server, which accounts round-trip delay between the edge and the server and all other request processing delays before hitting the inference model, seems to the the heavy part of the latency. 
%The large variation in the latency is due to the use of public cloud. 
%Note that this part of delay can be offset by deploying a GPU server locally. 
%It can be seen that the most expensive, as well as uncertain part is the communication delay, that depends on available network as we use public cloud. 
In our EARLIN-based setup, we improve this latency by not sending to the server when not required and thus getting rid of the communication delay. 
\newline
\textbf{Experimental Results: }
We show a set of aggregated results in Figure~\ref{acc_comp}. We show the accuracy results for varying degree of OOD samples for EARLIN, Baseline, and ``no detector''. We observe that as the OOD ratio $\rho$ rises, the accuracy drops sharply if no OOD detector is applied. The overall accuracy of Baseline also declines whereas the accuracy of EARLIN grows as the OOD ratio grows. This is because EARLIN has considerably higher TNR value and higher detection accuracy than the Baseline detector. It is to note that when $\rho$ is close to $0$ (very few samples are OOD compared to ID), the accuracy of EARLIN is slightly worse than that of when no detector is used. This is because EARLIN detects, in the worst case, 5\% ID samples as OODs (since TPR is 95\%), which contributes to reducing the overall accuracy. 

Figure~\ref{acccomp} shows the performance of EARLIN as we increase $\rho$. We observe that, as we increase $\rho$, the overall accuracy increases and the inference latency decreases. The decline in inference latency is due to the fact that as more OOD inputs are fed, they are detected at the edge as OOD. The samples being detected as ID are uploaded to the cloud accounting all three components of delay and the number of those samples decline as $\rho$ grows. We note that time required per inference when model is not associated with any detector is equivalent to the case when OOD samples are detected at the last layer of the model, as in both cases input will be sent to the server for classification and OOD detection.  In Figure~\ref{latcomp}, we show the average time required per inference in this case. In Section~\ref{collinf}, we showed that the overall accuracy of a model increases at a rate of $TNR - {acc_{\mathcal M}} \times TPR$ with the increase of $\rho$. Figure~\ref{acc_comp} shows how well that characterization fits with the experimental results. As we can see, our obtained curve closely matches the linear curve for the expected accuracy obtained based on our formulation. The same is true for cost (latency). We see that inference latency decreases linearly at a rate of $(T_C+T_S) \times (FPR-TPR)$, as expected.

\section{Conclusion and Future Works}
In this paper, we propose a novel edge-cloud collaborative inference system, EARLIN, based on a proposed Out-of-Distribution (OOD) detection technique. EARLIN enables the detection of OOD samples using feature maps obtained from the shallow layers of the pretrained deep learning classifiers. We exploit the advantage of early detection to design at OOD-aware edge-cloud collaborative inference framework as we deploy the small foot-print detector part on an edge device and the full model in the cloud. During inference, the edge detects if an input sample is ID. If it is, the sample is sent to the cloud for classification. Otherwise, the sample is reported as OOD and the edge starts processing the next sample in the pipeline. In this way, we make the inference at the edge faster and more precise. We characterize the performance and cost of the setup. Experimental results on benchmark datasets show that EARLIN performs well on OOD detection. Moreover, when deployed on a prototype implementation, results obtained show that expected improvement in cost and performance is achieved using proposed EARLIN-based setup. In future, we plan to investigate more on building a context-aware adaptive OOD detection setup that takes advantage of choosing from multiple candidate OOD detectors based on desired cost-accuracy trade offs.

%\begin{spacing}{0.7}

\bibliographystyle{splncs04}

%\end{spacing}

\end{document}